%% file: main.tex
\documentclass{article}

      \usepackage[nonatbib, final]{neurips_2021}

\usepackage[utf8]{inputenc} 
\usepackage[T1]{fontenc}    
\usepackage[colorlinks=true,citecolor=blue, linkcolor=magenta, urlcolor=blue]{hyperref}
\usepackage{url}            
\usepackage{booktabs}       
\usepackage{amsfonts}       
\usepackage{nicefrac}       
\usepackage{microtype}      

\usepackage{graphicx}
\usepackage{chngcntr}
\usepackage{wrapfig}
\usepackage{subcaption}
\usepackage{enumitem}
\usepackage[sort, numbers]{natbib}
\usepackage{comment}
\usepackage{caption}

\input{defs}

\title{Neural Additive Models: \\Interpretable Machine Learning with Neural Nets}

\author{
  Rishabh Agarwal$^{\thanks{Correspondence to: Rishabh Agarwal <\texttt{rishabhagarwal@google.com}>, Levi Melnick <\texttt{lemeln@microsoft.com}>, and Rich Caruana <\texttt{rcaruana@microsoft.com}>. }}$\\
  Google Research, Brain Team\\
  \And
  Levi Melnick\\
  Microsoft Research
  \AND
  Nicholas Frosst\\
  Cohere \\
  \And
  Xuezhou Zhang\\
  University of Wisconsin-Madison \\
  \And
  Ben Lengerich \\
  MIT
  \AND
  Rich Caruana\\
  Microsoft Research
  \And
  Geoffrey E. Hinton\\
  Google Research, Brain Team \\
}

\begin{document}

\maketitle

\vspace*{-0.2cm}
\begin{abstract}
\vspace*{-0.05cm}

Deep neural networks~(DNNs) are powerful black-box predictors that have achieved impressive performance on a wide variety of tasks. However, their accuracy comes at the cost of intelligibility: it is usually unclear how they make their decisions. This hinders their applicability to high stakes decision-making domains such as healthcare. We propose Neural Additive Models~(NAMs) which combine some of the expressivity of DNNs with the inherent intelligibility of generalized additive models. NAMs learn a linear combination of neural networks that each attend to a single input feature. These networks are trained jointly and can learn arbitrarily complex relationships between their input feature and the output. Our experiments on regression and classification datasets show that NAMs are more accurate than widely used intelligible models such as logistic regression and shallow decision trees. They perform similarly to existing state-of-the-art generalized additive models in accuracy, but are more flexible because they are based on neural nets instead of boosted trees. To demonstrate this, we show how NAMs can be used for multitask learning on synthetic data and on the COMPAS recidivism data due to their composability, and demonstrate that the differentiability of NAMs allows them to train more complex interpretable models for COVID-19.  Source code is available at \href{https://neural-additive-models.github.io}{\ttfamily neural-additive-models.github.io}. 


\end{abstract}


\vspace*{-0.2cm}
\section{Introduction}
While deep neural networks have achieved impressive results on tasks such as computer vision~\citep{he2016deep} and language modeling~\citep{radford2019language}, it is notoriously difficult to understand how such networks make predictions, and they are often considered as black-box models. This hinders their applicability to high-stakes domains such as healthcare, finance and criminal justice.  Various efforts have been made to demystify the predictions of neural networks~(NNs). For example, one family of methods, represented by LIME~\citep{ribeiro2016should}, attempt to \emph{explain} individual predictions of a neural network by approximating it locally with interpretable models such as linear models and shallow trees\footnote{Linear models, shallow decision trees and GAMs are interpretable only if the features they are trained on are interpretable.}. However, these approaches often fail to provide a global view of the model and their explanations often are not faithful to what the original model computes or do not provide enough detail to understand the model's behavior~\citep{rudin2019stop}. 

In this paper, we make restrictions on the \emph{structure} of neural networks, which yields a family of glass-box models called Neural Additive Models~(NAMs), that are inherently interpretable while suffering little loss in prediction accuracy when applied to tabular data. Methodologically, NAMs belong to a model family called Generalized Additive Models~(GAMs)~\citep{hastie1990generalized}. 
GAMs have the form:
\begin{equation}
    g(\expected[y]) = \beta + f_1(x_1) + f_2(x_2) + \dots + f_K(x_K) \label{eq:gam}
    \vspace{-0.2cm}
\end{equation}

\begin{wrapfigure}{r}{0.53\linewidth}
    \centering
    \vspace{-0.35cm}
    \includegraphics[width=\linewidth]{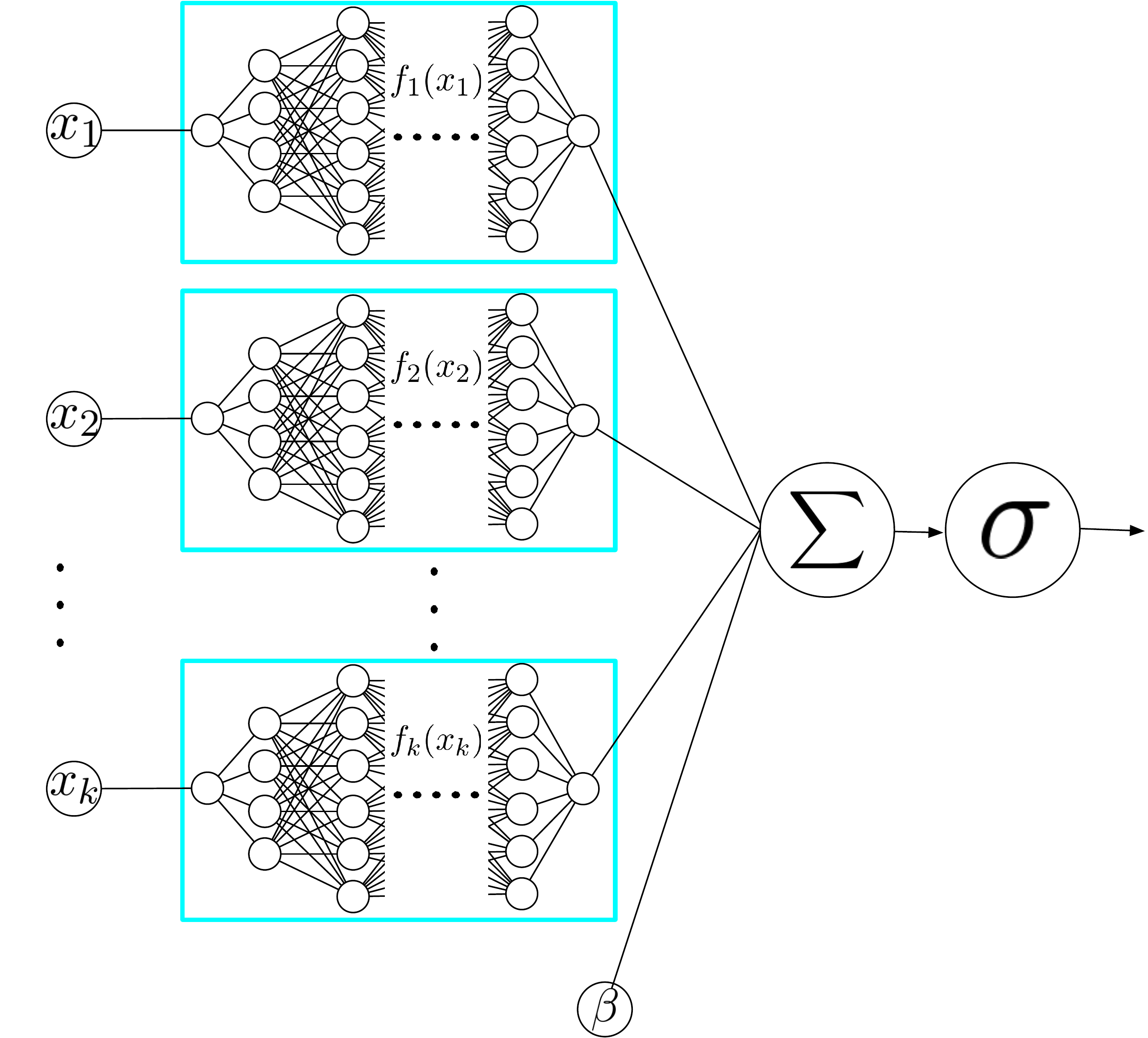}
    \vspace{-0.5cm}
    \caption{NAM architecture for binary classification. Each input variable is handled by a different neural network. This results in easily interpretable yet highly accurate models.}\label{fig:NAM_model_sigmoid}
    \vspace{-0.6cm}
\end{wrapfigure}

where $\rvx = (x_1,\ x_2,\ \dots,\ x_K)$ is the input with $K$ features, $y$ is the target variable, $g(.)$ is the link function (\eg logistic function) and each $f_i$ is a univariate shape function with $\expected[f_i]=0$. Generalized linear models, such as logistic regression, are a special form of GAMs where each $f_i$ is restricted to be linear.

NAMs learn a linear combination of networks that each attend to a single input feature: each $f_i$ in \eqref{eq:gam} is parametrized by a neural network. These networks are trained jointly using backpropagation and can learn arbitrarily complex shape functions. 
Interpreting NAMs is easy as the impact of a feature on the prediction does not rely on the other features and can be understood by visualizing its corresponding shape function~(\eg plotting $f_i(x_i)$ \versus $x_i$). While interpretability of NAMs may seem heuristic, the graphs learned by NAMs are an exact description of how NAMs compute a prediction.


Traditionally, GAMs were fitted via iterative backfitting using smooth low-order splines, which reduce overfitting and can be fit analytically. More recently, GAMs~\citep{caruana2015intelligible} were fitted with boosted decision trees to improve accuracy and to allow GAMs to learn jumps in the feature shaping functions to better match patterns seen in real data that smooth splines could not easily capture.  This paper examines using DNNs to fit generalized additive models~(NAMs) which provides the following advantages:
\begin{itemize}[topsep=1pt, partopsep=1pt, leftmargin=15pt, parsep=0pt, itemsep=6pt]
    \item NAMs 
    introduce an expressive yet intelligible class of models to the deep learning~(DL) community, a much larger community than the one using tree-based GAMs. 
    \item NAMs are likely to be combined with other DL methods in ways we don't foresee. This is important because a key drawback of deep learning is interpretability. For example, NAMs have already been employed for survival analysis~\citep{utkin2021survnam}.
    \item NAMs, due to the flexibility of NNs, can be easily extended to various settings problematic for boosted decision trees. For example, extending boosted tree GAMs to multitask, multi-class or multi-label learning requires significant changes to how trees are trained, but is easily accomplished with NAMs without requiring changes to how neural nets are trained
    due to their composability~(\Secref{sec:multitask_nams}). Futhermore, the differentiability of NAMs allows them to train more complex interpretable models for COVID-19~(\Secref{sec:intelligble_param_gen}).
    \item Graphs learned by NAMs are not just an explanation but an exact description of how NAMs compute a prediction. As such, a decision-maker can easily interpret NAMs and understand exactly how they make decisions. This would help harness the expressivity of neural nets on high-stakes domains with intelligibility requirements, \eg in-hospital mortality prediction~\citep{lee2021development}. 
    \item NAMs are more scalable as inference and training can be done on GPUs/TPUs or other specialized hardware using the same toolkits developed for deep learning over the past decade -- GAMs currently cannot. 
    \item Accurate GAMs~\citep{caruana2015intelligible} currently require millions of decision trees to fit each shape function while NAMs only use a small ensemble~(2 - 100) of neural nets. Thus, NAMs are relatively much easier to extend compared to GAMs.

  \end{itemize}
  

\vspace{-0.1cm}
\section{Neural Additive Models}
\label{sec:new_activation}
\vspace{-0.1cm}

\begin{figure*}[t]
  \begin{center}
    \footnotesize
    \begin{tabular}{@{}c@{}c@{}}
      \includegraphics[width=0.46\linewidth]{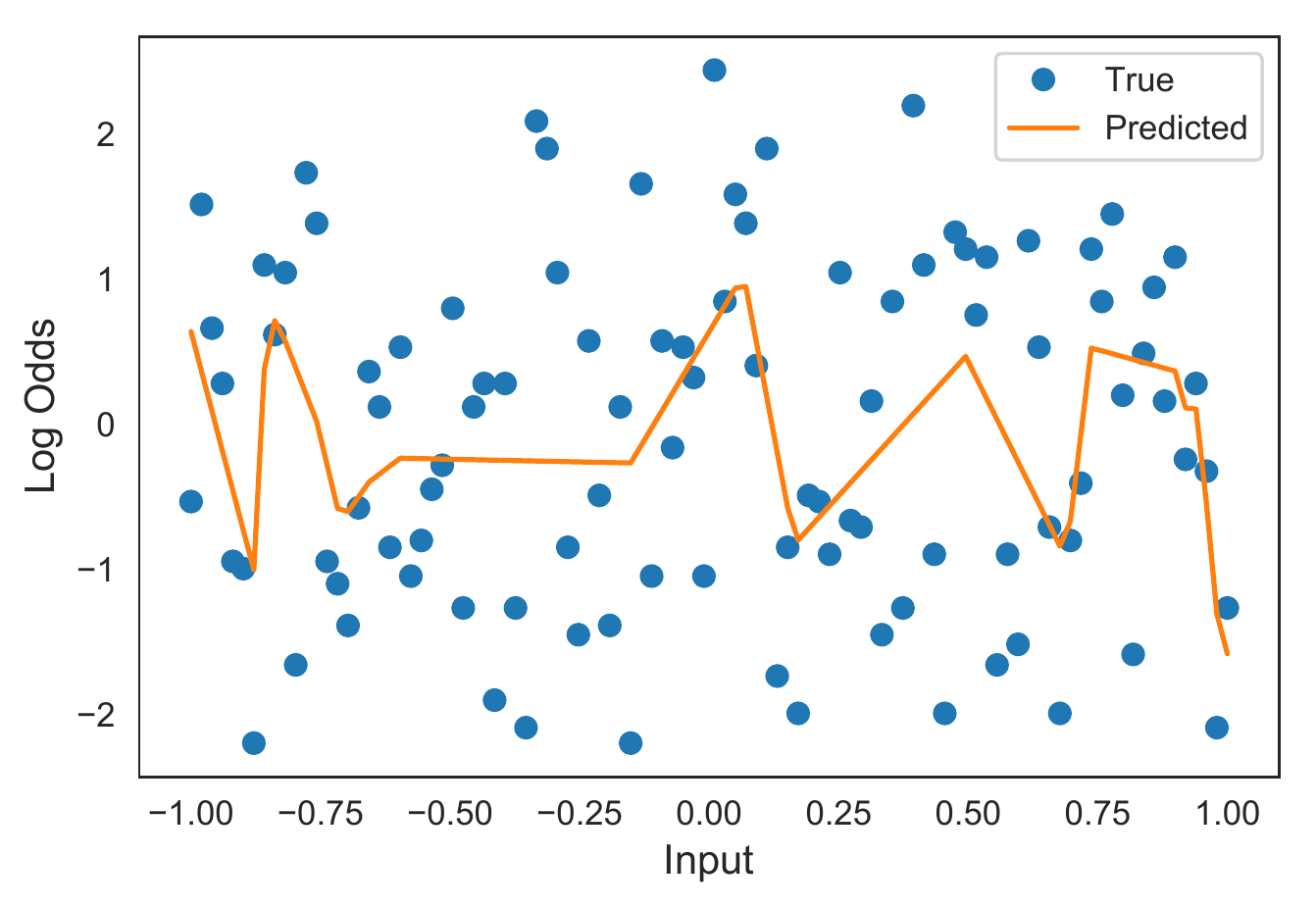} & \includegraphics[width=0.46\linewidth]{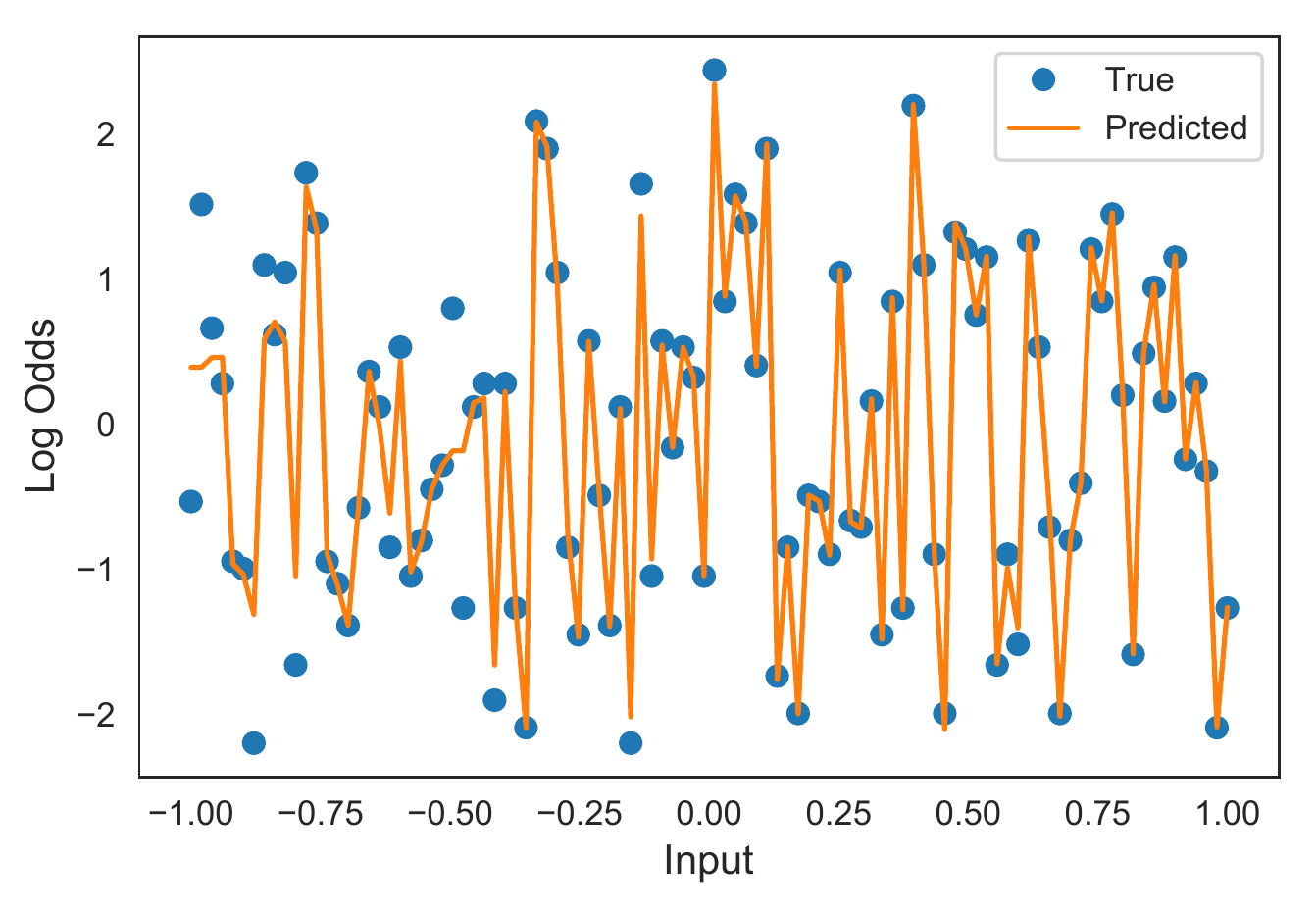} \\
      (a) & (b)
    \end{tabular}
    \vspace{-0.1cm}
    \caption{{\bf Accurately Fitting the Toy Dataset}: Training predictions learned by a single hidden layer neural network with 1024 (a) standard ReLU, and (b) ReLU-$n$ with ExU hidden units trained for 10,000 epochs on the binary classification dataset described in Section~\ref{sec:new_activation}. We can see that the ReLU network has learned a fairly smooth function while the ExU network has learned a very jumpy function. We find that a DNN with three hidden layers also learned smooth functions~(see Figure~\ref{fig:dnn_toy}).}\label{fig:fit_relus}
  \end{center}
\vspace{-0.6cm}
\end{figure*}

\begin{figure}[t]
\begin{minipage}{0.44\textwidth}
        \centering
    \includegraphics[width=0.96\linewidth]{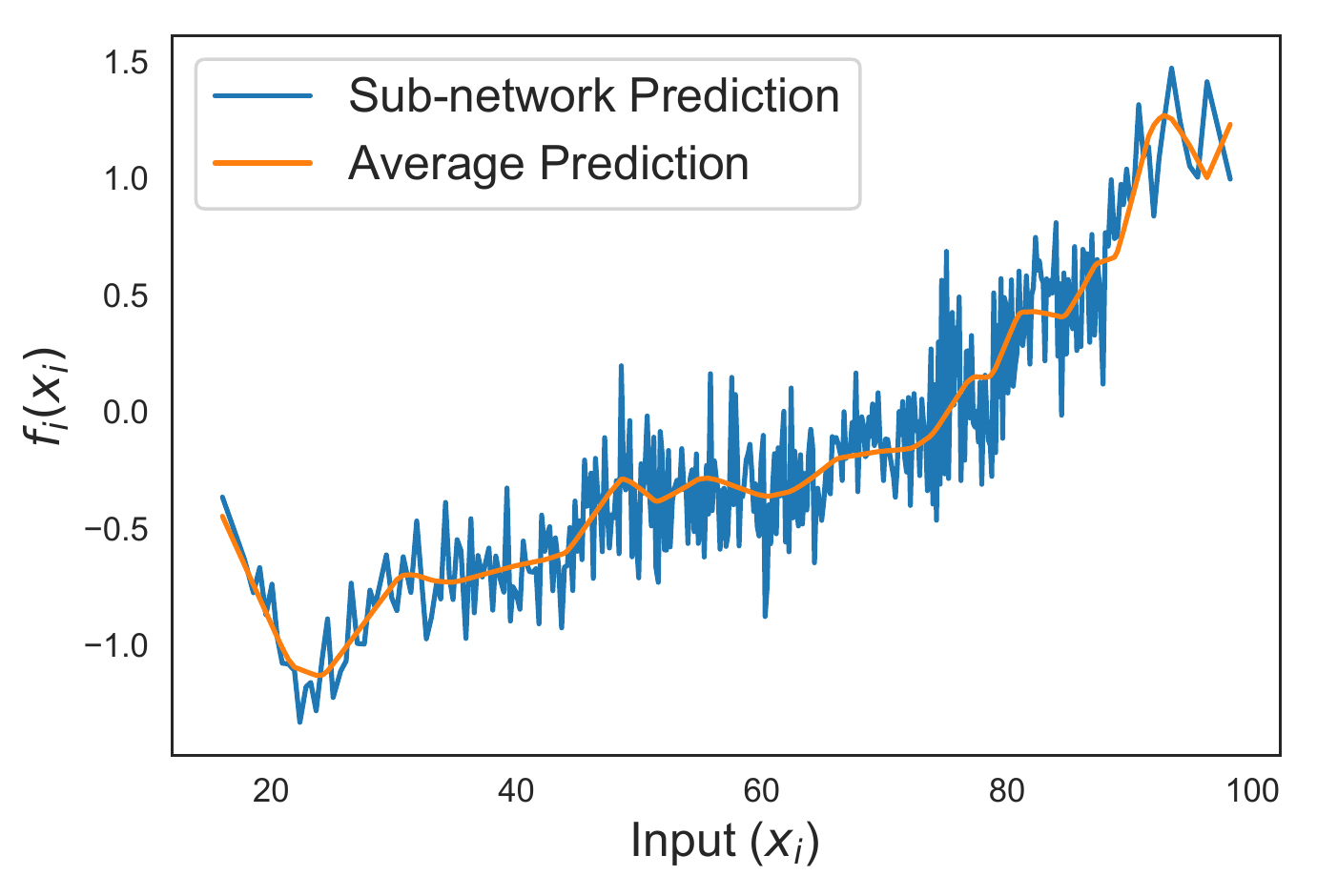}
    \caption{{\bf Regularizing ExU networks.} Output of a ExU feature net trained with dropout = $0.2$ for the age feature in the MIMIC-II dataset~\citep{saeed2011multiparameter}. Predictions from individual subnets~(as a result of dropping out hidden units) are much more jagged than the average predictions using the entire feature net. Refer to Section~\ref{sec:regularization} for an overview of regularization approaches used in this work. }\label{fig:dropout_vs_without}
    \vspace{-0.25cm}
\end{minipage}~~~~~
\begin{minipage}{0.55\textwidth}

    \captionsetup[subfigure]{aboveskip=-0.4pt, belowskip=-0.4pt}
    \centering
    \begin{subfigure}{0.9\textwidth}
      \includegraphics[width=\textwidth]{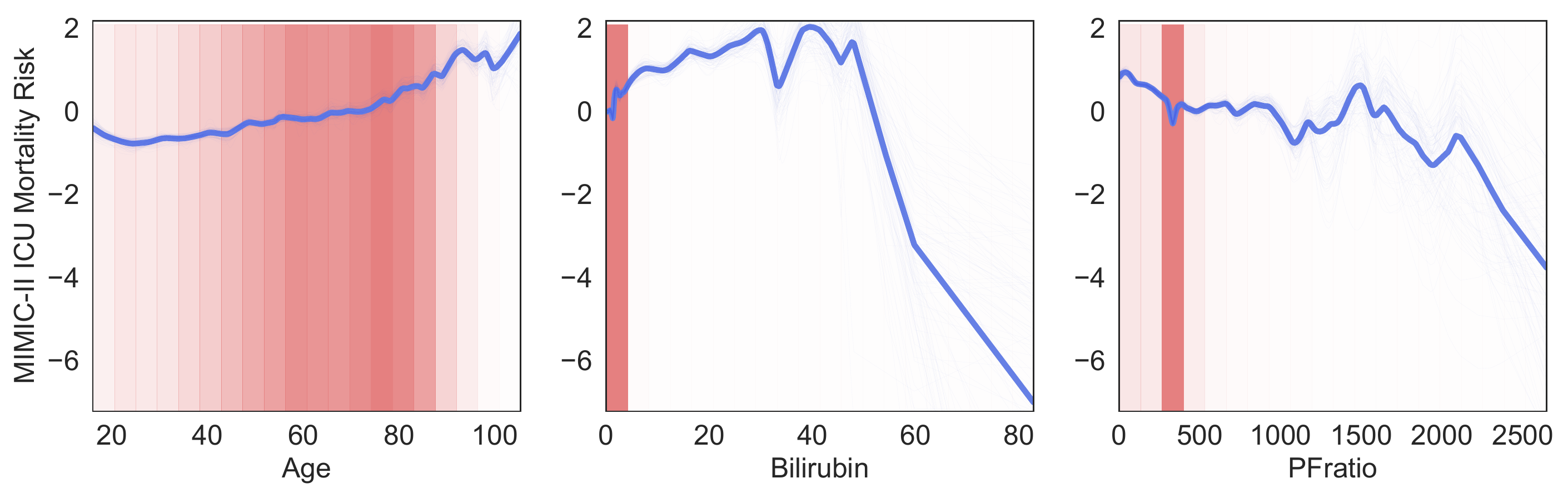}
      \caption{Graphs learned by NAMs with ExU units}
      \label{fig:Ng1}
    \end{subfigure}
    \vspace{0.1cm}
    \begin{subfigure}{0.9\textwidth}
      \includegraphics[width=\textwidth]{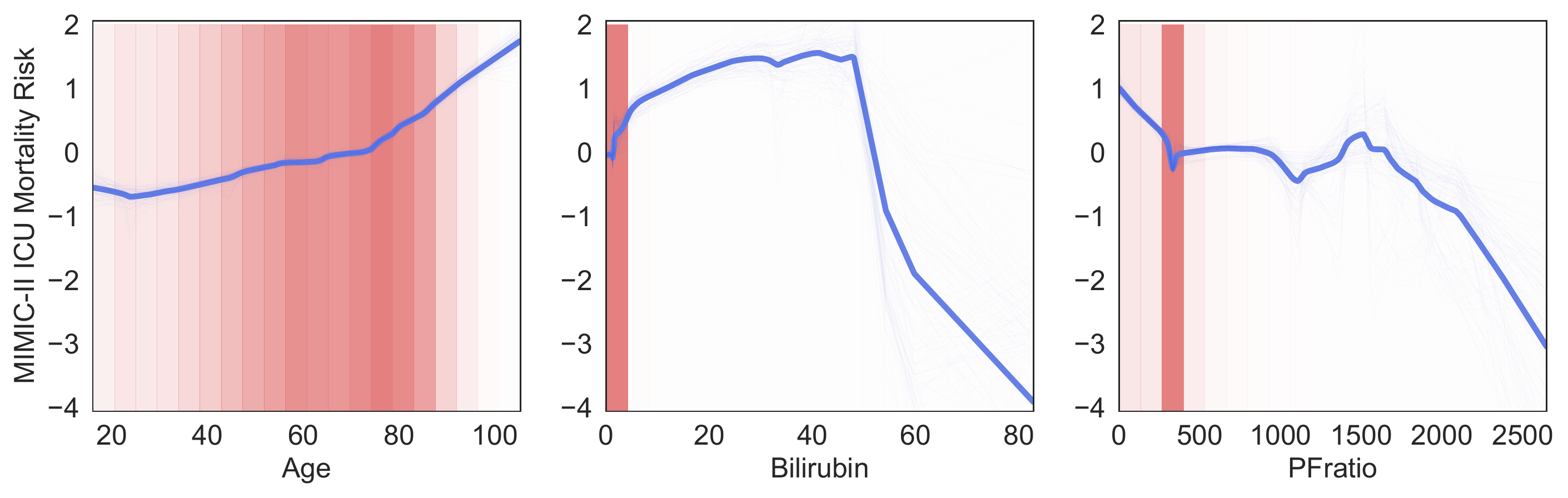}
      \caption{Graphs learned by NAMs with standard units}
      \label{fig:Ng2}
    \end{subfigure}%
    \vspace{-0.05cm}
    \caption{{\bf ExU \versus standard} hidden units. On MIMIC-II, NAMs trained with ExU units learn jumpier graphs than with standard units while achieving a similar AUC~($\approx 0.829$). Ensembling them further improves performance~($\approx 0.830$). Note that white regions in the plots correspond to regions with low data density (typically a few points) and thus we see much higher variance in the learned shape functions. We present a detailed case study on the MIMIC-II dataset in \Secref{sec:mimic2}.}\label{fig:jumps}
    \vspace{-0.5cm}
\end{minipage}
\end{figure}

\textbf{Modeling jagged shape functions} is required to learn accurate additive models as there are often sharp jumps in real-world datasets, \eg see Figure~\ref{fig:jumps} for jumps in graphs for PFRatio and Bilirubin which correspond to real patterns in the MIMIC-II dataset~\citep{saeed2011multiparameter}~(Section~\ref{sec:mimic2}). Similarly, \citet{caruana2015intelligible} observe that GAMs fit using splines tend to over regularize and miss genuine details in real data, yielding less accuracy than tree-based GAMs. Therefore, we require that neural networks~(NNs) are able to learn highly non-linear shape functions, to fit these patterns.

Although NNs can approximate arbitrarily complex functions~\citep{hornik1989multilayer}, we find that standard NNs fail to model highly jumpy 1D functions, and demonstrate this failure empirically using \textbf{a toy dataset}. 
The toy dataset is constructed as follows: 
For the input  $x$, we sample 100 evenly spaced points in [-1, 1].
For each $x$, we sample $p$ uniformly random in [0.1, 0.9) and generate 100 labels from a Bernoulli random variable which takes 
the value 1 with probability $p$. This creates a binary classification dataset of $(x, y)$ tuples with 10,000 points. 
Figure~\ref{fig:fit_relus} shows the log-odds of the empirical probability $p$~(\ie $\log \frac{p}{1-p}$) of classifying the label of $x$ as 1 for each input $x$. This dataset tests the NN's ability to ``overfit'' the data, rather than its ability to generalize.

Over-parameterized NNs with ReLUs~\citep{nair2010rectified} and standard initializations such as Kaiming initialization~\citep{he2015delving} and Xavier initialization~\citep{glorot2010understanding} struggle to overfit this dataset when trained using mini-batch gradient descent, despite the NN architecture being expressive enough\footnote{This problem doesn't occur with full-batch gradient descent.}(see Figures~\ref{fig:fit_relus}(a) and \ref{fig:dnn_toy}). This difficulty of learning large local fluctuations with ReLU networks without affecting their global behavior when fitting jagged functions might be due to their bias towards smoothness~\citep{rahaman2018spectral, arpit2017closer}.

We propose \textit{exp-centered}~(ExU) hidden units to overcome this neural net failure: we simply learn the weights in the logarithmic space with inputs shifted by a bias. Specifically, for a scalar input $x$, each hidden unit using an activation function $f$ computes $h(x)$ given by
\begin{equation}
    h(x) = f\left(e^{w} * (x - b)\right) \label{eq:exu_unit}
\end{equation}
where $w$ and $b$ are the weight and bias parameters. The intuition behind ExU units is as follows: For modeling jagged functions, a hidden unit should be able to change its output significantly, with a tiny change in input. This requires the unit to have extremely large weight values depending on the sharpness of the jump. The ExU unit computes a linear function of input where the slope can be very steep with small weights, making it easier to modify the output easily during training. ExU units do not improve the expressivity of neural nets, however they do improve their learnability for fitting jumpy functions. While we use ExU units to train accurate NAMs, they are more generally applicable for approximating jumpy functions with neural nets.


We noticed that ExU units with standard weight initialization also struggle to learn 
jagged curves; instead initializing the weights using a normal distribution $\mathcal{N}(x, 0.5)$ with $x \in [3, 4]$ works well in practice. This initialization simply ensures that the initial network starts with a jagged (random) function which we empirically find to be crucial for fitting any jumpy function. Furthermore, we use ReLU activations capped at $n$~(ReLU-$n$)~\citep{krizhevsky2010convolutional} to ensure that each ExU unit is active in a small input range, making it easier to model sharp jumps in a function without significantly affecting the global behavior. ExU-units can be combined with any activation function (\ie any $f$ can be used in~\eqref{eq:exu_unit}), but ReLU-$n$ performs well in practice. Figure~\ref{fig:fit_relus}(b) shows that NNs with ExU units are able to fit the toy dataset significantly better than standard NNs.

Finally, realistic shape functions typically tend to be smooth with large jumps at only a few points~(\figref{fig:jumps}). To avoid overfitting with ExUs, strong regularization is crucial which can learn such realistic functions~(\eg \figref{fig:dropout_vs_without}). With ReLUs, we can typically fit smooth functions but they might miss some of these jumps. To avoid overfitting when fitting NAMs with ExUs, we employ various regularization methods  including dropout, weight decay, output penalty, and feature dropout~(see Section~\ref{sec:regularization} for an overview).

\vspace{-0.2cm}
\subsection{Intelligibility and Modularity of NAMs}
\vspace{-0.2cm}

    
    

The intelligibility of NAMs results in part from the ease with which they can be visualized. Because each feature is handled independently by a learned shape function parameterized by a neural net, one can get a full view of the model by simply graphing the individual shape functions. For data with a small number of inputs, it is possible to have an accessible explanation of the model's behavior visualized fully on a single page. Please note these shape function plots are not just an explanation but an \textit{exact} description of how NAMs compute a prediction. A decision-maker can easily interpret such models and understand exactly how they make decisions, for example, we validated the behavior of NAMs on the MIMIC-II dataset~\citep{saeed2011multiparameter} with a doctor~(Appendix~\ref{sec:mimic2}).


We set the average score for each graph~(\ie each feature) averaged across the entire training dataset to zero by subtracting the mean score. To make individual shape functions identifiable and modular,
a single bias term is then added to the model so that the average predictions across all data points matches the observed baseline. This makes interpreting the contribution of each term easier:~\eg on binary classification tasks, negative scores decrease probability, and positive scores increase probability compared to the baseline probability of observing that class. This property also allows each graph to be removed from the NAM~(zeroed out) without introducing bias to the predictions. 

{\bf Visualization}. We plot each shape function and the corresponding data density on the same graph. Specifically, we plot each learned shape function $f_k(x_k)$ \versus $x_k$ for an ensemble of NAMs using a semi transparent blue line, which allows us to see when the models in the ensemble learned the same shape function and when they diverged. This provides a sense of the confidence of the learned shape functions. We also plot on the same graphs the normalized data density, in the form of pink bars. The darker the shade of pink, the more data there is in that region. This allows us to know when the model had adequate training data to learn appropriate shape functions. 

\section{Evaluating the Accuracy of NAMs}
\vspace{-0.2cm}

\begin{table}[t]
    \centering
    \caption{\textbf{Single-task learning NAM results}. Means and standard deviations are reported from 5-fold cross validation.
    Higher AUCs and lower RMSEs are better. We report results on two widely used regression datasets, namely California Housing~\citep{pace1997sparse} for predicting housing prices and FICO~\citep{fico} for understanding credit score predictions, as well as two classification datasets, namely Credit~\citep{dal2015adaptive} for financial fraud detection and MIMIC-II~\citep{saeed2011multiparameter} for predicting mortality in ICUs. We present a case study on the MIMIC-II dataset in \Secref{sec:mimic2} and discuss the interpretations from NAMs on other datasets in \Secref{sec:additional_datasets}.}
    \vspace{0.2cm}
    \label{table:classification}
    \begin{tabular}{| c | c | c | c | c |} 
     \hline
     Model & MIMIC-II (AUC) & Credit (AUC) & CA Housing (RMSE) & FICO (RMSE)  \\ [0.5ex] 
     \hline\hline
     Log./Linear Reg. &  0.791 $\pm$ 0.007 & 0.975 $\pm$ 0.010 & 0.728 $\pm$ 0.015 & 4.344 $\pm$ 0.056\\ 
     CART &  0.768 $\pm$ 0.008&0.956 $\pm$ 0.004 & 0.720 $\pm$ 0.006 & 4.900 $\pm$ 0.113\\
     \hline\hline
     NAMs &  0.830 $\pm$ 0.008 & 0.980 $\pm$ 0.002 & 0.562 $\pm$ 0.007 &  3.490 $\pm$ 0.081 \\
     EBMs &  0.835 $\pm$ 0.007 & 0.976 $\pm$ 0.009 & 0.557 $\pm$ 0.009 & 3.512 $\pm$ 0.095\\
     \hline\hline
     XGBoost &  0.844 $\pm$ 0.006 & 0.981 $\pm$ 0.008 & 0.532 $\pm$ 0.014 & 3.345 $\pm$ 0.071\\
     DNNs &  0.832 $\pm$ 0.009 & 0.978 $\pm$ 0.003 & 0.492 $\pm$ 0.009 & 3.324 $\pm$ 0.092\\
     \hline
    \end{tabular}
\end{table}

\begin{figure*}[t]
    \centering
    \begin{minipage}{0.66\textwidth}
        \centering
      \includegraphics[width=\linewidth]{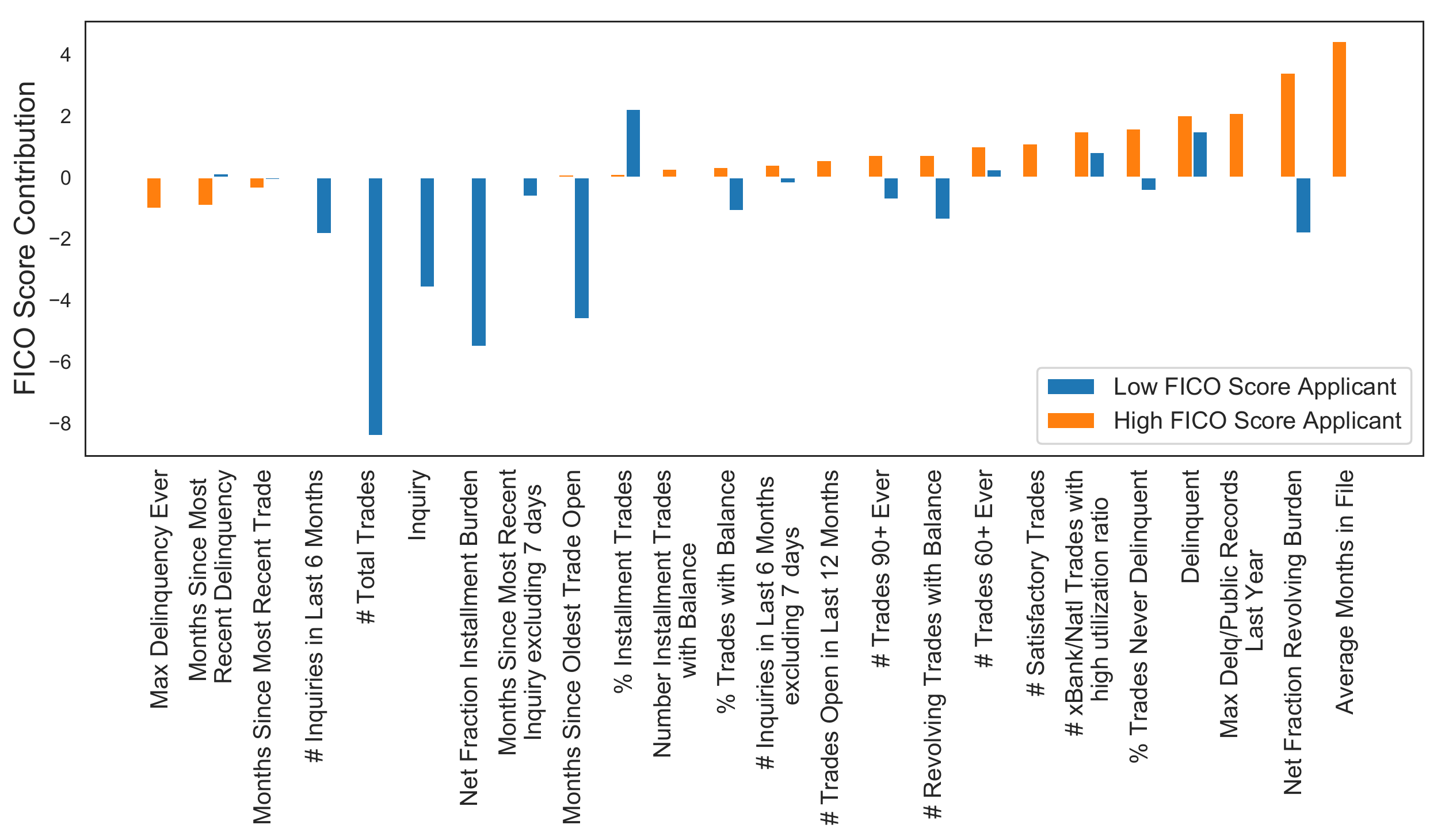}
      \vspace{-0.1cm}
      \caption{{\bf Understanding individual predictions for credit scores}. Feature contribution using the learned NAMs for predicting scores of two applicants in the FICO dataset~\citep{fico}. For a given input, each feature net in the NAM acts as a lookup table and returns a contribution term. These contributions are combined in a modular way: they are added up, and passed through a link function for prediction. the longer a person's credit history, the better it is for their credit score
      The high scoring applicant has a long credit history~(Average Months on File), which contributes to their credit score better.  On the contrary, the low scoring applicant used their credit quite frequently~(Total Number of Trades) and has a large burden~(Net Fraction Installment Burden), thus resulting in a low score.}\label{fig:fico_local}
    \end{minipage}\hfill
    \begin{minipage}{0.3\textwidth}
        \centering
        \vspace{-0.3cm}
        \includegraphics[width=0.9\textwidth]{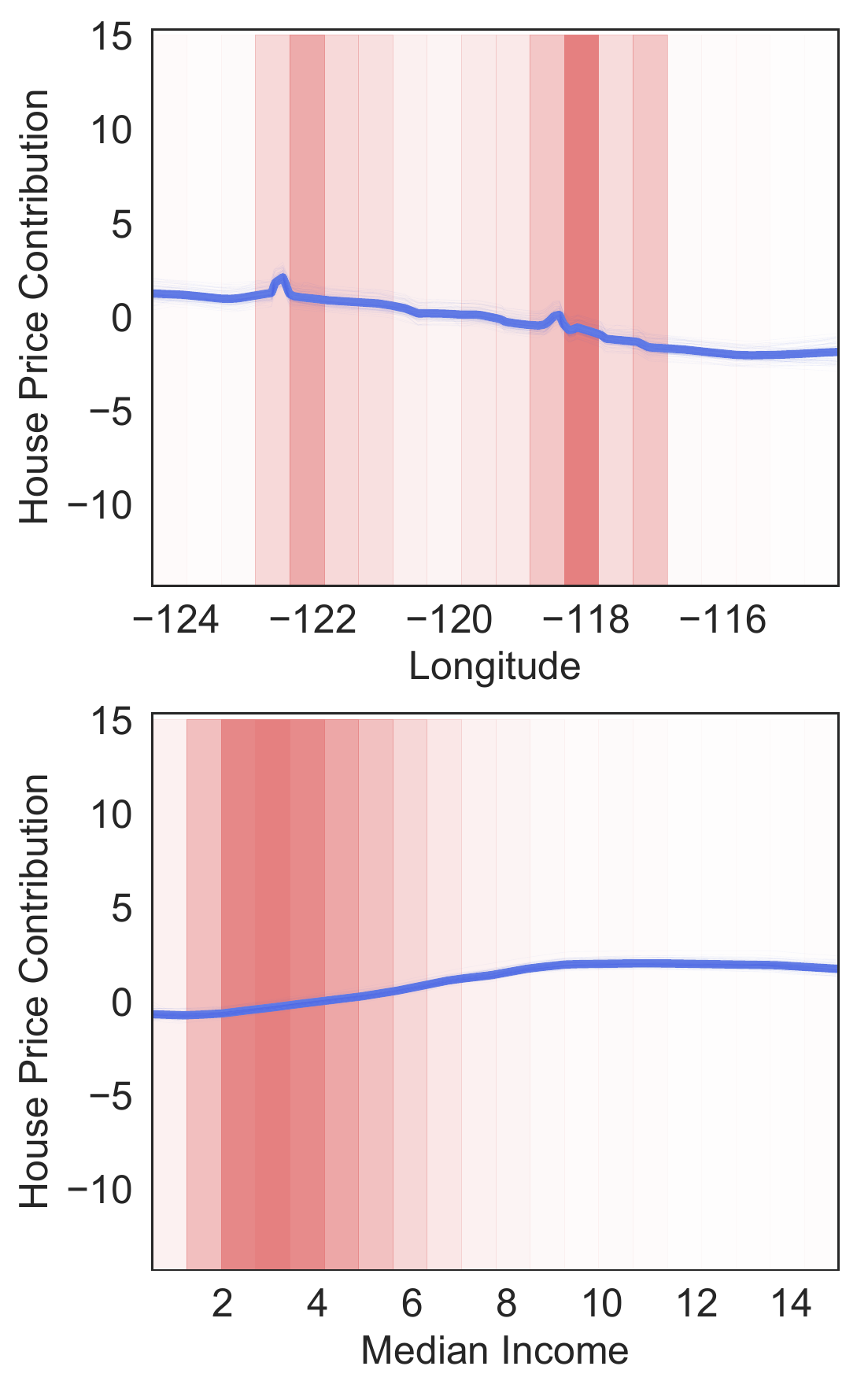} 
        \vspace{-0.1cm}
        \caption{{\bf California Housing}. Graphs learned by NAMs trained to predict house prices~\citep{pace1997sparse} for two most important features. As expected, The house prices increase linearly with median income in high data density regions. Furthermore, the graph for longitude shows sharp jumps in price prediction around the location of San Francisco and Los Angeles.}\label{fig:housing_paper}
    \end{minipage}
    \vspace{-0.55cm}
\end{figure*}

\label{sec:experiments}
In this section, we evaluate the single-task learning capacity of NAMs against the following baselines on both regression and classification tasks:
\begin{itemize}[topsep=0pt, partopsep=0pt, leftmargin=13pt, parsep=0pt, itemsep=4pt]
    \item{\bf Logistic / Linear Regression and Decision Trees (CART)}: Prevalent intelligible models. For both methods above we use the \texttt{sklearn} implementation~\citep{pedregosa2011scikit}, and tune the hyper-parameters with grid search.
    \item {\bf Explainable Boosting Machines~(EBMs)}: Current state-of-the-art GAMs~\citep{caruana2015intelligible, lou2012intelligible} which use gradient boosting of millions of shallow bagged trees that cycle one-at-a-time through the features. 
    \item {\bf Deep Neural Networks~(DNNs)}: Unrestricted, full-complexity models which can model higher-order interaction between the input features. This gives us a sense of how much accuracy we sacrifice in order to gain interpretability with NAMs.
    \item {\bf Gradient Boosted Trees (XGBoost)}: Another class of full-complexity models that provides an upper bound on the achievable test accuracy in our experiments. We use the XGBoost implementation~\citep{chen2016xgboost}.
\end{itemize}

{\bf Training and Evaluation}. Feature nets in NAMs are selected amongst (1) DNNs containing 3 hidden layers with 64, 64 and 32 units and ReLU activation, and (2) single hidden layer NNs with 1024 ExU units and ReLU-$1$ activation. We perform 5-fold cross validation to evaluate the accuracy of the learned models. To measure performance in \tabref{table:classification},  we use area under the precision-recall curve~(AUC) for binary classification and root mean-squared error~(RMSE) for regression. More details about training and evaluation protocols can be found in Section~\ref{sec:train} in the appendix. 

NAMs achieve comparable performance to EBMs on both classification and regression datasets, making them a competitive alternative to EBMs. Given this observation, we next look at some additional capabilities of NAMs that are not available to EBMs or any tree-based learning methods.

\vspace{-0.4cm}
\section{Unique Capabilities of NAMs}
\vspace{-0.25cm}

\subsection{Intelligible Parameter Generation: Leveraging the Differentiability of NAMs}
\label{sec:intelligble_param_gen}
\begin{figure*}[t]
    \centering
    \begin{subfigure}[c]{0.4\textwidth}
        \centering
        \fbox{
        \includegraphics[width=0.92\textwidth]{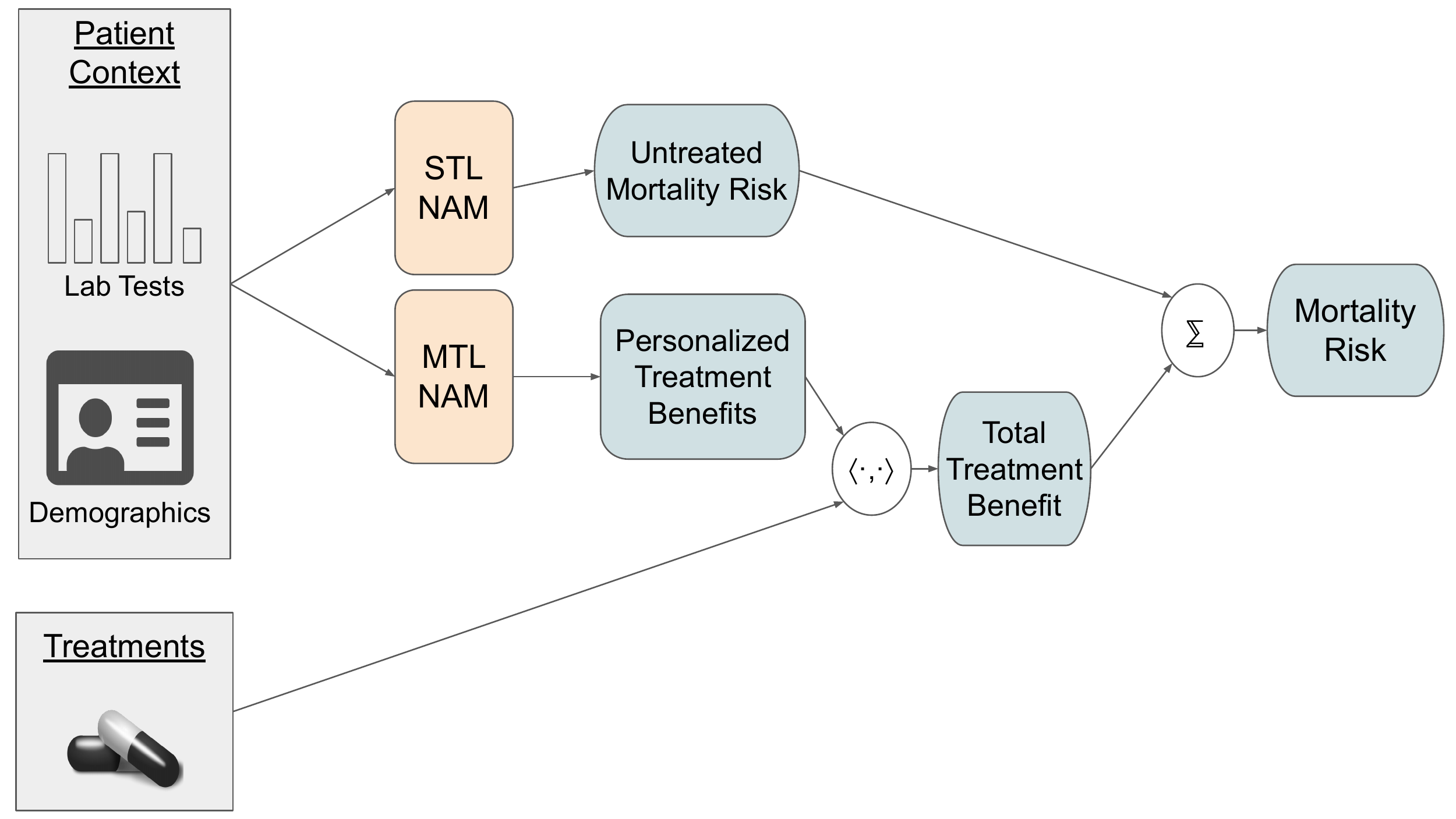}
        }
        \caption{\small{Architecture}}
    \end{subfigure}
    \begin{subfigure}[c]{0.18\textwidth}
        \centering
        \begin{subfigure}[b]{\textwidth}
            \includegraphics[width=\textwidth]{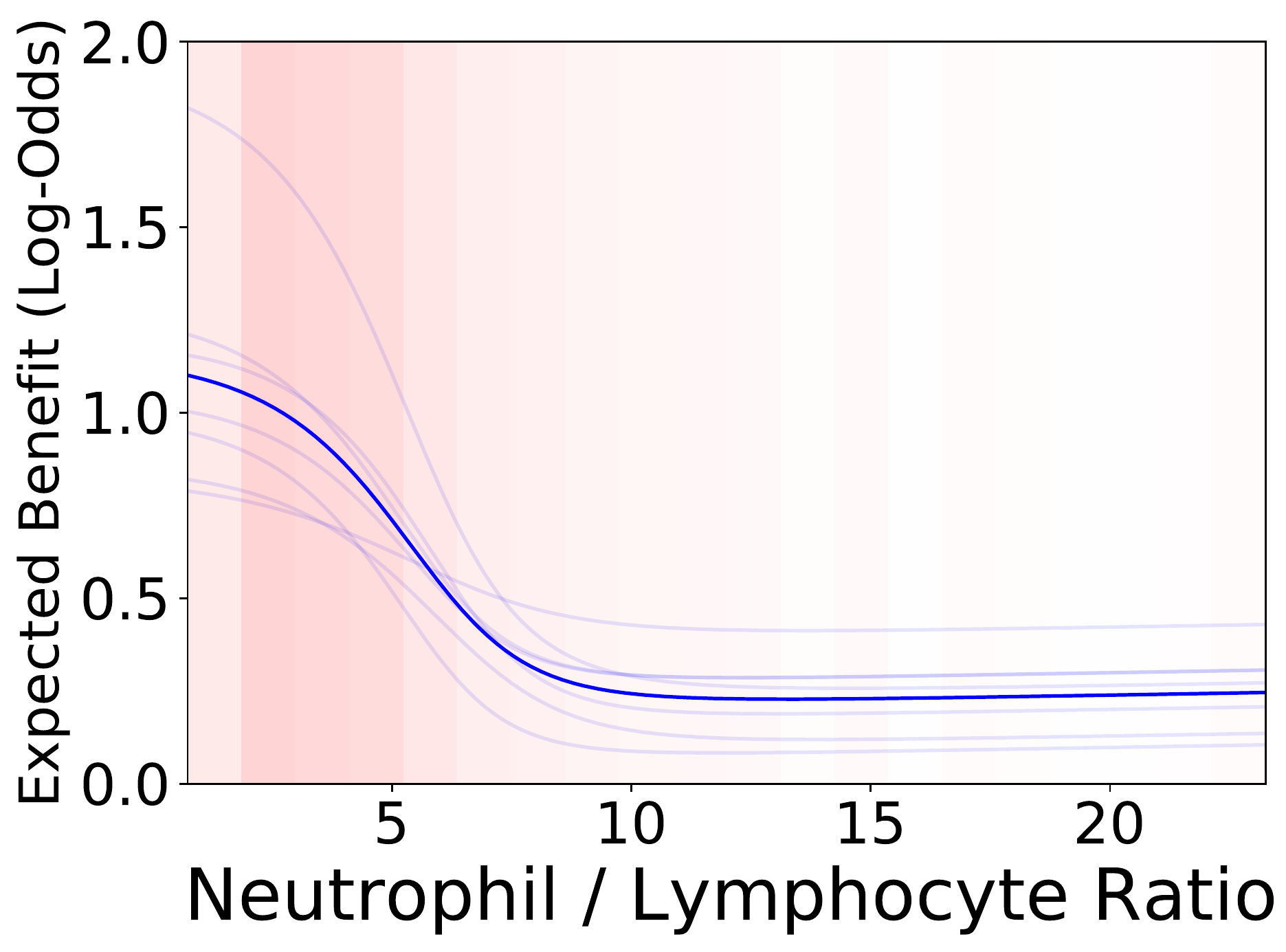}
            \includegraphics[width=\textwidth]{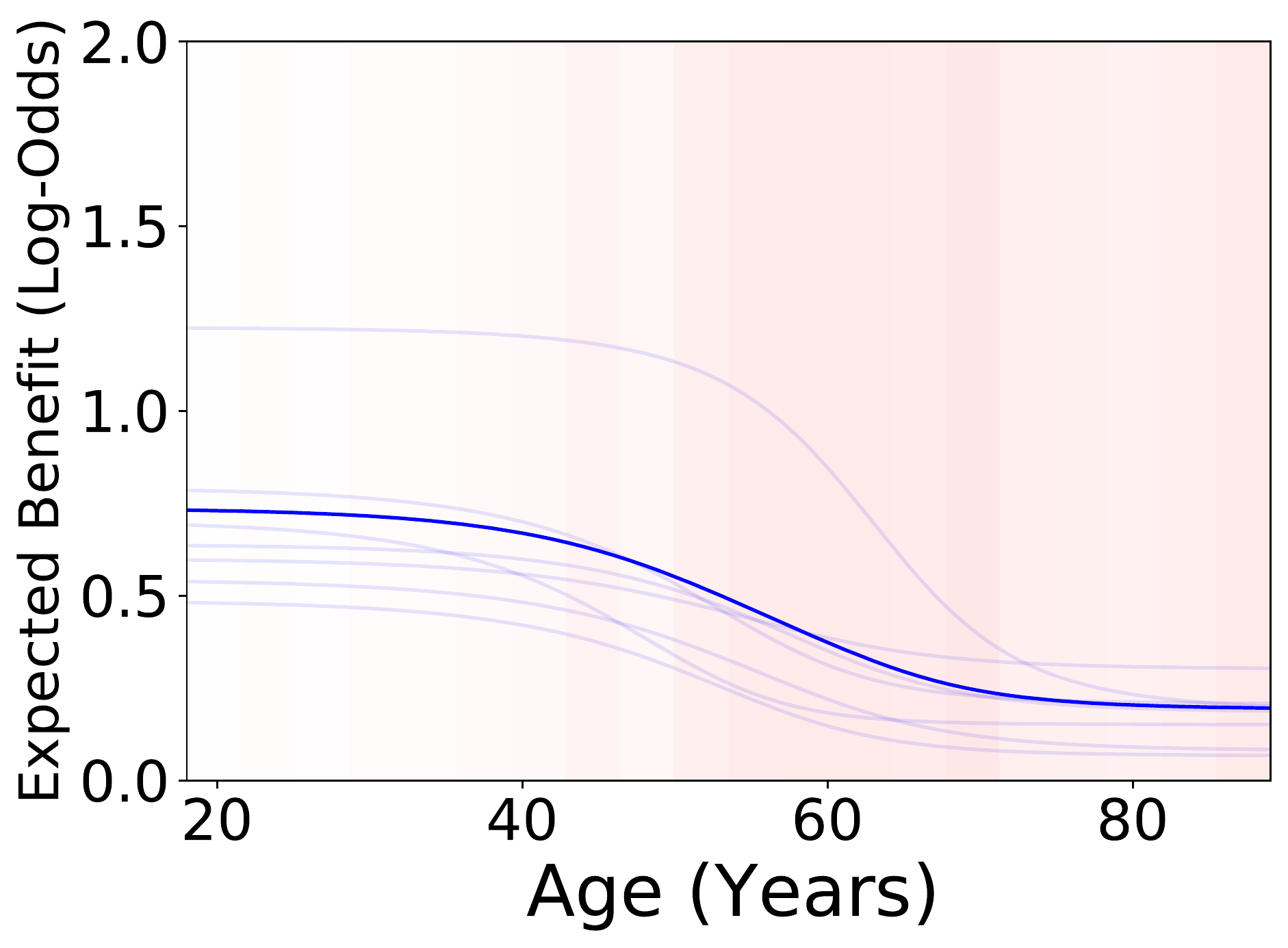}
        \end{subfigure}
        \caption{\footnotesize{Anti-Coagulants}}
    \end{subfigure}
    ~
    \begin{subfigure}[c]{0.18\textwidth}
        \centering
        \begin{subfigure}[b]{\textwidth}
            \includegraphics[width=\textwidth]{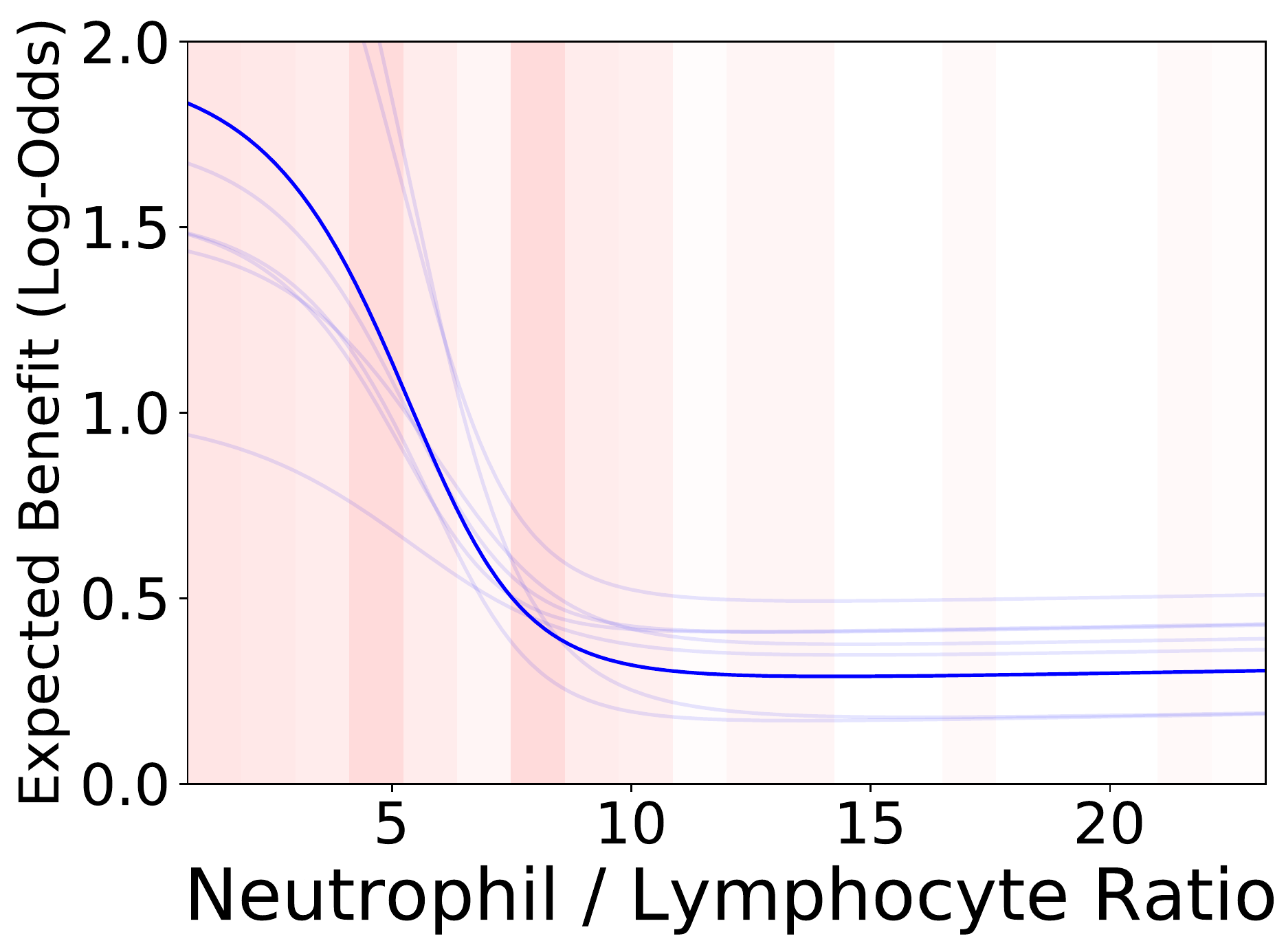}
            \includegraphics[width=\textwidth]{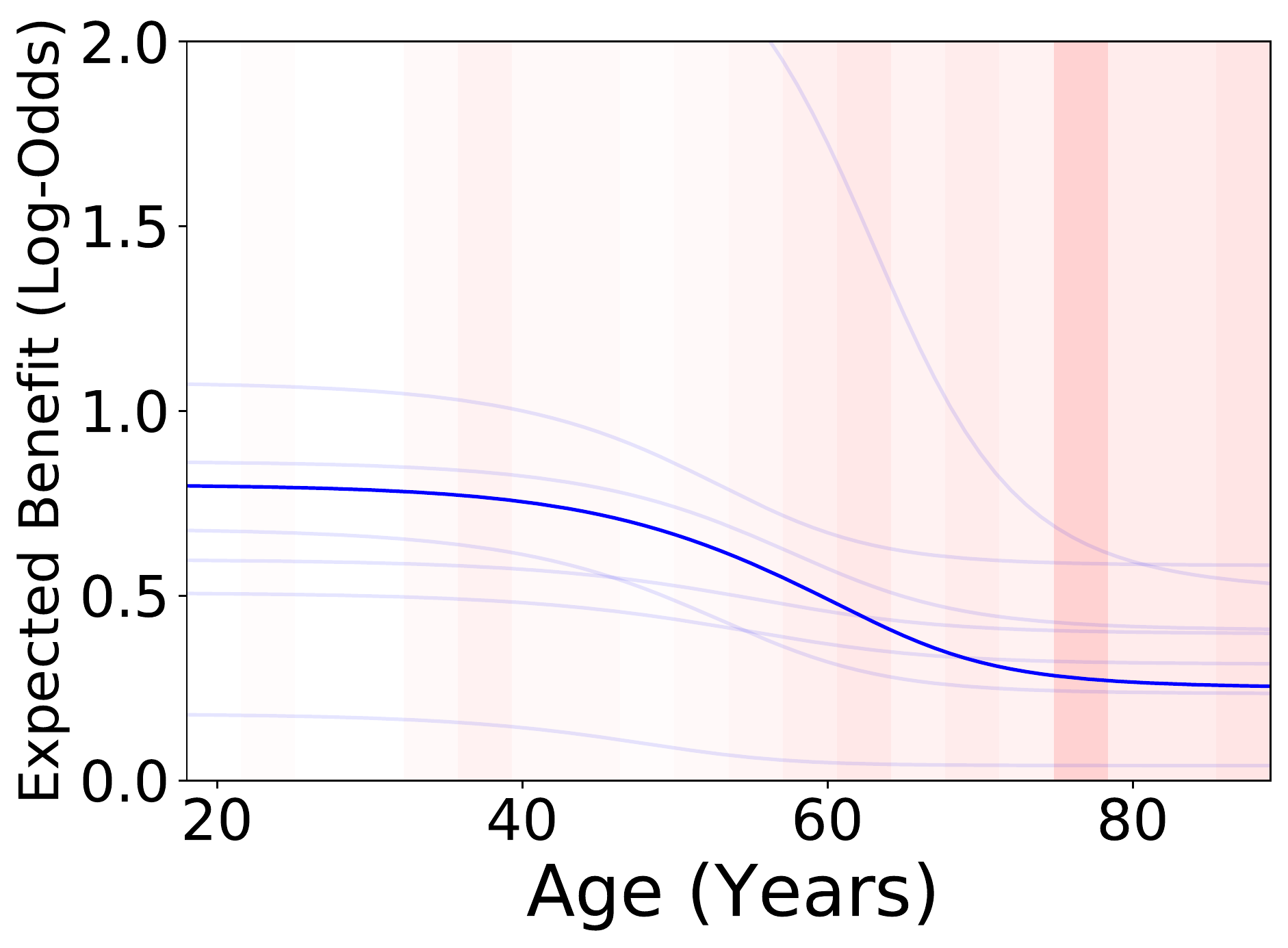}
        \end{subfigure}
        \caption{\small{NSAIDs}}
    \end{subfigure}
    ~
    \begin{subfigure}[c]{0.18\textwidth}
        \centering
        \begin{subfigure}[b]{\textwidth}
            \includegraphics[width=\textwidth]{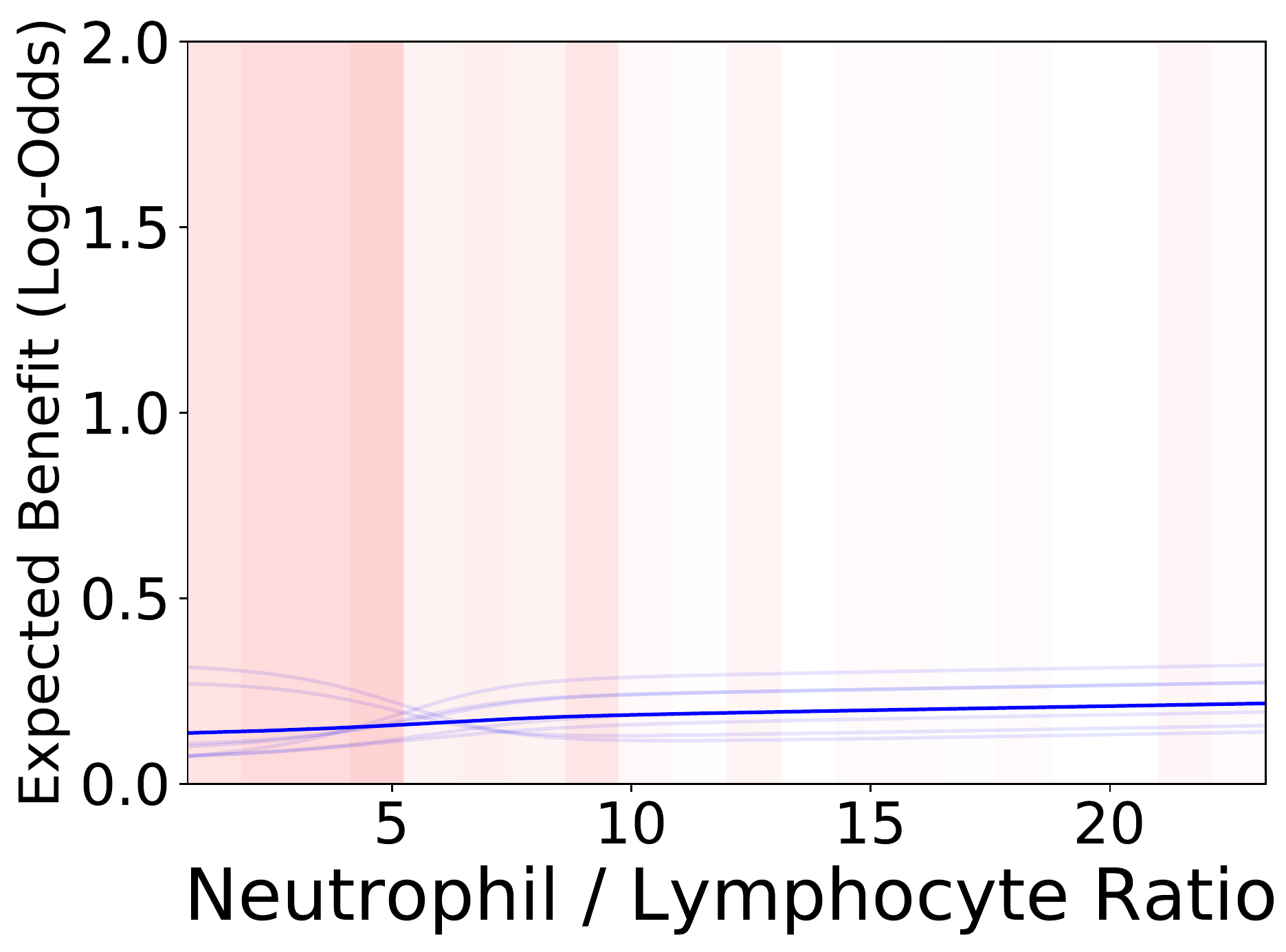}
            \includegraphics[width=\textwidth]{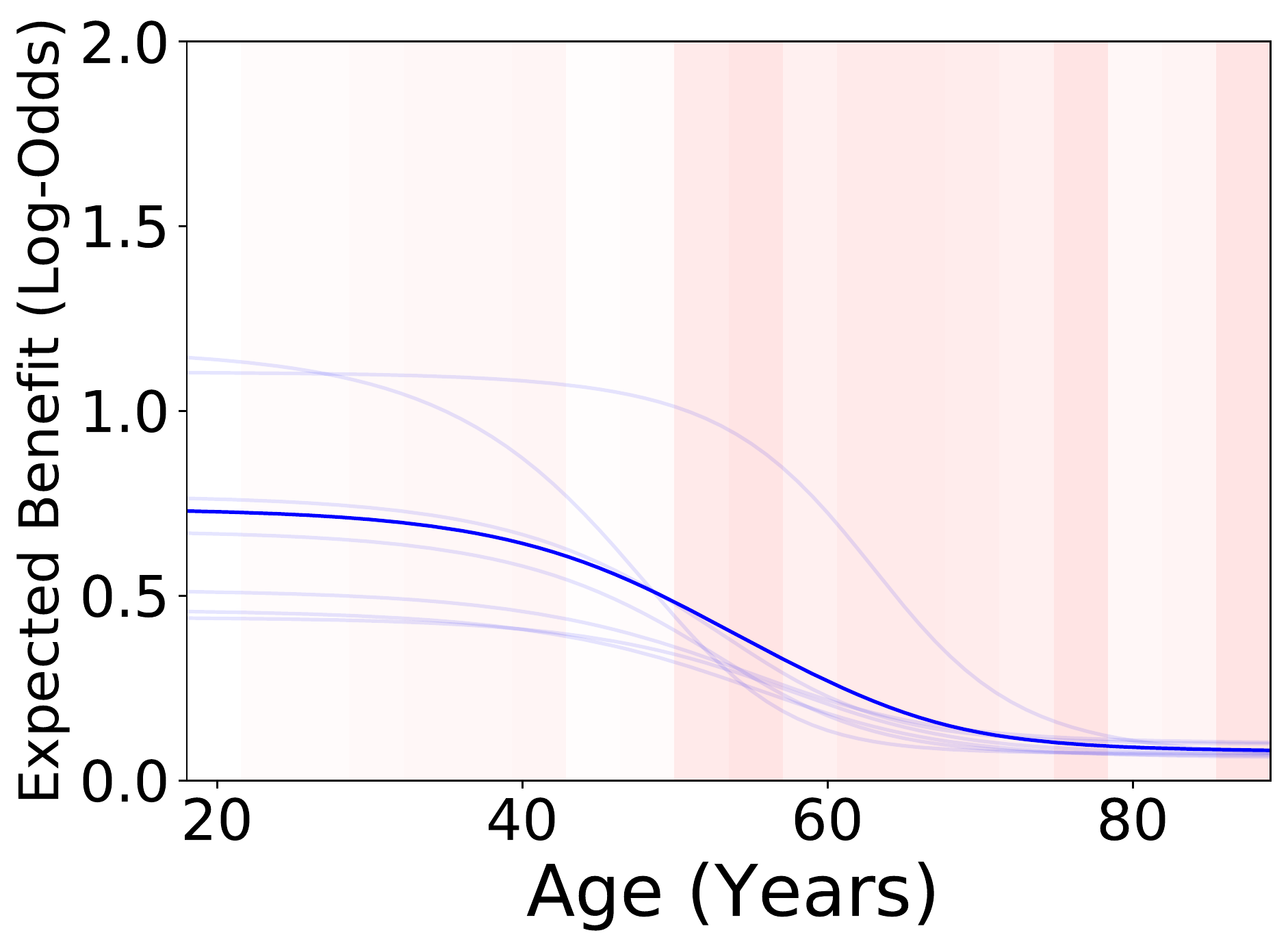}
        \end{subfigure}
        \caption{\small{Glucocorticoids}}
    \end{subfigure}
    \caption{{\bf Estimating personalized treatment benefits for Covid-19 patients.}
    NAMs provide a unique combination of intelligibility and differentiability which make them suitable as a component in contextual parameter generation (a). By applying NAMs in this way, we are able to estimate and interpret personalized benefits of medical treatments for Covid-19 patients (b-d).} 
    \label{fig:param_generation}
    \vspace{-0.5cm}
\end{figure*}

Medical treatment protocols are designed to deliver treatments to patients who would most benefit from them. 
To optimize treatment protocols, we would like a model which provides an intelligible map from patient information to an estimate of benefit for each potential treatment. 
To accomplish this, we use a NAM to generate parameters for personalized models of mortality risk given treatment (Fig.~\ref{fig:param_generation}). 
By training to match predicted mortality risk with observed mortality, the NAM encodes expected treatment benefits as a function of patient information. 
NAMs are the only nonlinear GAM suitable for this application because NAMs are differentiable and can be trained via backpropagation.

Figure~\ref{fig:param_generation} shows a NAM trained to predict treatment benefits for Covid-19 patients. 
We train the model on deidentified data from over 3000 Covid-19 patients.  The model suggests that the benefits of anti-coagulants and NSAIDs decrease with increased Neutrophil / Lymphocyte Ratio (NLR), while the effectiveness of glucocorticoids slightly increases with increasing NLR. 
NLR is a marker of inflammation and severe Covid-19; it is thus expected that anti-coagulants (which target a distinct biomedical pathway) and NSAIDs (which are weaker) would not be as effective for patients with elevated NLR. 
In contrast, glucocorticoids become more effective for patients with more inflammation. 
This example shows the utility of a \emph{differentiable} nonlinear additive model such as NAMs.

\vspace{-0.2cm}
\subsection{Multitask Learning}\label{sec:multitask_nams}
\vspace{-0.1cm}

\begin{wrapfigure}{r}{0.56\linewidth}
    \centering
    \includegraphics[width=\linewidth]{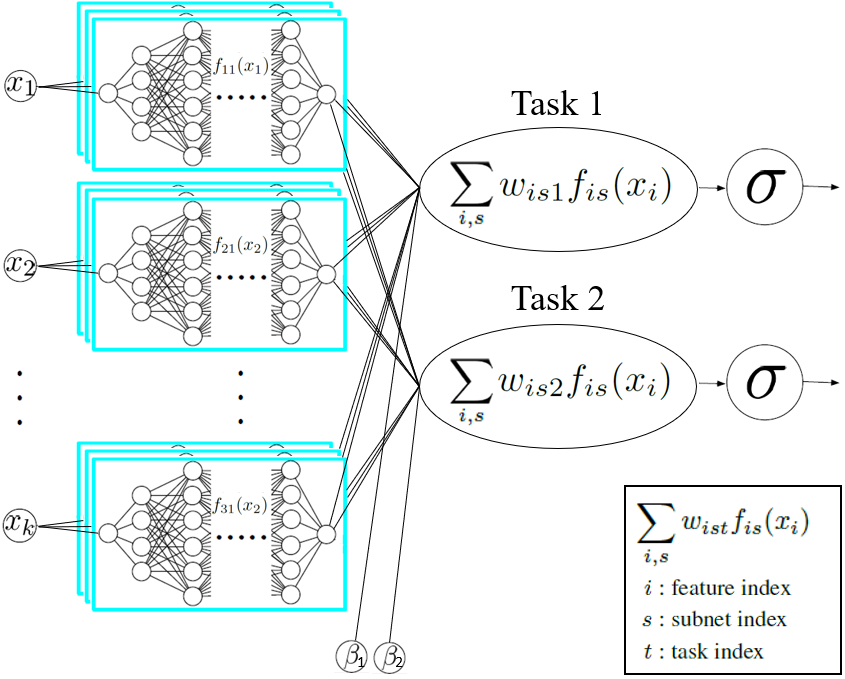}
    \vspace{-0.5cm}
    \caption{\textbf{Multitask NAM architecture} for binary classification. Multiple subnets are trained on each input feature and weighted sums are learned over the subnets.}\label{fig:NAM_model_sigmoid_mtl}
    \vspace{-0.5cm}
\end{wrapfigure}

One advantage of NAMs is that they are easily extended to multitask learning~(MTL)~\citep{caruana1997mtl}, whereas MTL is not available in EBMs or in any major boosted-tree package. In NAMs, the composability of neural nets makes it easy to train multiple subnets per feature. The model can learn task-specific weights over these subnets to allow sharing of subnets (shape functions) across tasks while also allowing subnets to differentiate between tasks as needed. However, it is unclear how to implement MTL in EBMs and possibly requires changes to both the backfitting procedure and the information gain rule in decision trees. Figure~\ref{fig:NAM_model_sigmoid_mtl} shows a multitask NAM architecture that can jointly learn different feature representations for each task while preserving the intelligibility and modularity of NAMs. As we show, this can benefit both accuracy and interpretability. We first demonstrate multitask NAMs on a synthetic dataset before showing their utility on a multitask formulation of the COMPAS recidivism prediction dataset.

\textbf{Multitask NAM Architecture}. The multitask architecture is identical to that of single task NAMs except that each feature is associated with multiple subnets and the model jointly learns a task-specific weighted sum over their outputs that determines the shape function for each feature and task. The outputs corresponding to each task are summed and a bias is added to obtain the final prediction score. The number of subnets does not need to be the same as the number of tasks --- the number of subnets can be less than, equal to, or even more than the number of tasks. Although the shape plot for each task is a linear combination of the shape plots learned by each subnet for that feature, this generates a single unique shape plot for each task and there is no need to examine what has been learned by the individual subnets for interpreting multitask NAMs. 

\subsubsection{Experiments on Synthetic Multitask Data}

Multitask models often show improvement over single task learning when tasks are similar to each other and training data is limited. 
We construct a synthetic dataset that showcases the benefit of multitask learning in NAMs and demonstrates their ability to learn task-specific shape plots when needed. We define 6 related tasks, each a function of three variables.  All 6 tasks are the same function of variables $x_0$ and $x_1$, and differ only in the function applied to $x_2$: 

\begin{table*}[h]
\centering
\begin{tabular}{l l l}
$Task_0 = f(x_0) + g(x_1) + h(x_2)$ & 
&
$Task_1 = f(x_0) + g(x_1) + i(x_2)$ \\
$Task_2 = f(x_0) + g(x_1) - h(x_2)$ &
&
$Task_3 = f(x_0) + g(x_1) - i(x_2)$ \\
$Task_4 = f(x_0) + g(x_1) + (h(x_2)+i(x_2))$ &
&
$Task_5 = f(x_0) + g(x_1) - (h(x_2)+i(x_2))$ \\
\end{tabular}
\end{table*}

Functions $f(x_0)$, $g(x_1)$, $h(x_2)$ and $i(x_2)$ are as follows:

\begin{figure}[h!]
    \vspace{-0.5cm}
    \centering
    \includegraphics[width=0.6\linewidth]{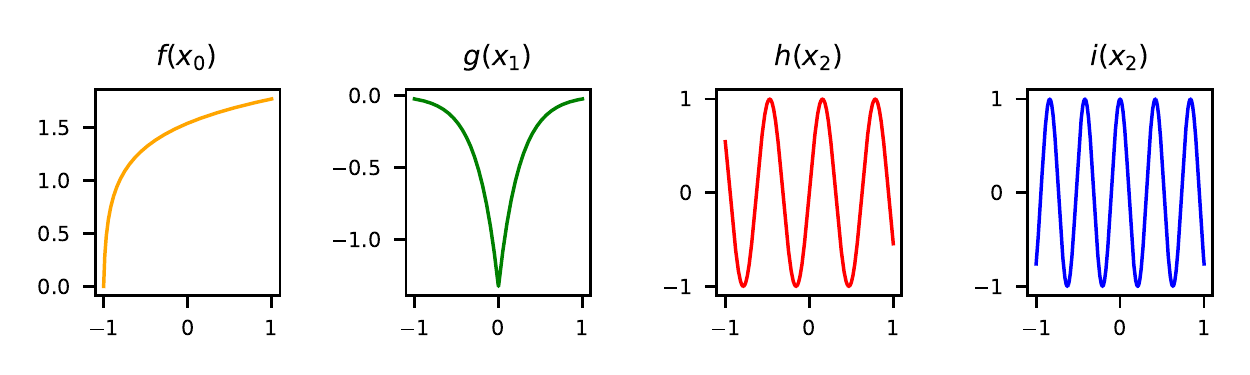}
    \label{fig:syntheticgenerators}
    \vspace{-0.5cm}
\end{figure}

\begin{figure*}[b!]
    \centering
    \includegraphics[trim={1cm 0 0 0}, width=1.1\linewidth]{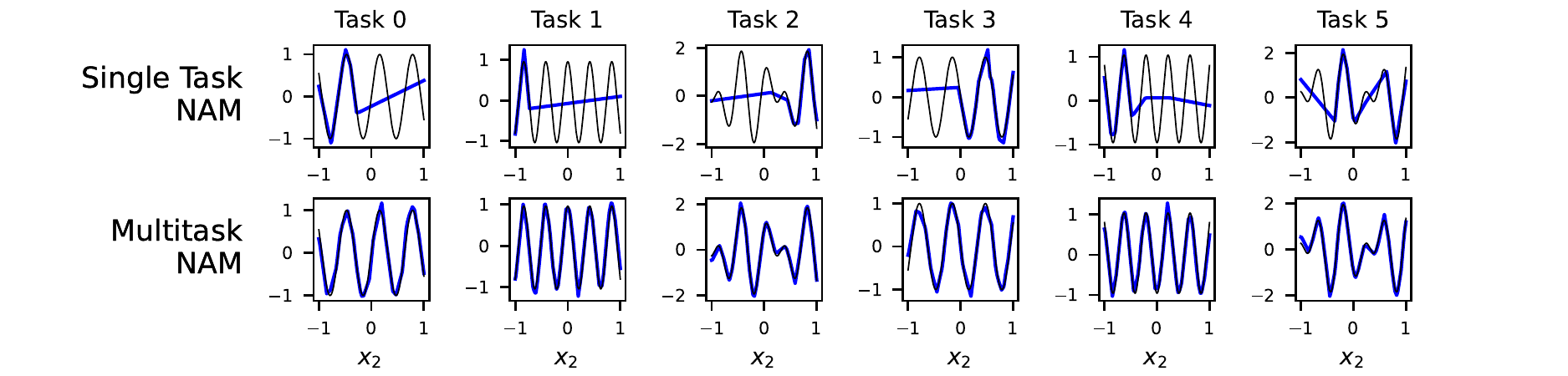}
    \vspace{-0.5cm}
    \caption{{\bf Single and Multitask NAM shape plots} for $x_2$ from a typical (median) run of each task. The learned shape function is blue; the generator function is black. See \ref{sec:synthetic_generators} for details of the generator functions.}
    \label{fig:syntheticx2}
    \vspace{-0.0cm}
\end{figure*}

\begin{table*}[h!]
\centering
\vspace{0.0cm}
\caption{MSE for STL and MTL NAMs on synthetic data. Average of 20 runs. Lower MSEs are better.}
\label{table:mtlsynthetic}
\begin{tabular}{| c | c | c | c | c | c | c | c |} 
 \hline
 Model & Task 0 & Task 1 & Task 2 & Task 3 & Task 4 & Task 5 & Mean \\ [0.5ex] 
 \hline\hline
 \quad Single Task NAM & 0.965 & 1.116 & 1.347 & 0.944 & 1.058 & 1.066 & 1.083\\ 
 \quad Multitask NAM & 0.710 & 0.715 & 0.709 & 0.711 & 0.717 & 0.709 & 0.712\\
\hline
\end{tabular}
\vspace{-0.25cm}
\end{table*}

\begin{figure}[t]
    \centering
    \includegraphics[trim={0 2cm 0 0}, width=1.0\linewidth]{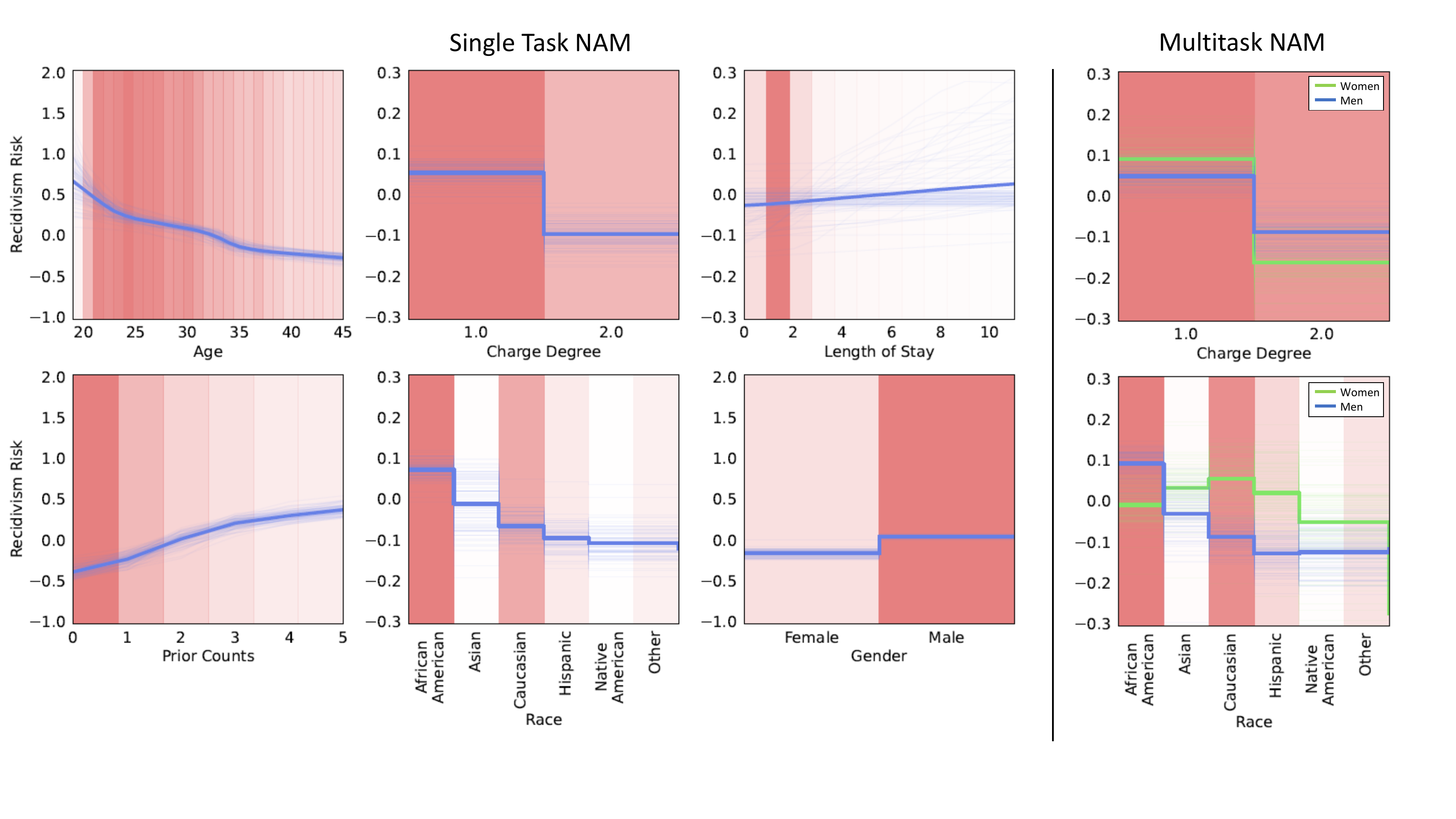}
    \vspace{-0.35cm}
    \caption{{\bf Single and Multitask COMPAS Recidivism Prediction}. Plots in the left column show the shape functions for each input feature learned by an ensemble of 100 single task NAMs. Thin blue lines represent shape functions for individual members of the ensemble. Pink bars represent the normalized data density for each feature. Plots in the right column show the Race and Charge degree shape plots for an ensemble of 100 multitask NAMS, with the Women task shown in green, and the Men task in blue.}
    \label{fig:recidivismmtl}
    \vspace{-0.5cm}
\end{figure}

A NAM with two subnets per feature can model every function of $x_2$ by learning two subnets, one for $h(x_2)$ and one for $i(x_2)$ and assigning appropriate weights to the output of each.  Because we would not know this in advance with real data, we use 6 subnets so that each of the 6 tasks (outputs) could, if needed, learn independent shape functions. We train models on 2,500 training examples, evaluate them on a test set of 10,000 examples, and average the results over 20 trials. Also, we ensured that each subnet has enough parameters to easily learn the necessary feature shape plots. So MTL is not doing better than STL because STL has inadequate capacity and MTL has more capacity.

Table~\ref{table:mtlsynthetic} shows that on average across all tasks, multitask NAMs achieve mean squared error 34\% lower than single task NAMs, and at least 25\% lower on each individual task. In all 120 trials of the 6 tasks combined, MTL achieved a better score than STL on 119 of the 120 trials. 
Figure~\ref{fig:syntheticx2} shows the shape plots learned by median runs of STL and MTL for the functions of $x_2$ that vary among tasks. Furthermore, we illustrate that a multi-task NAM is as interpretable as a single task NAM by plotting the multi-task NAM predictions on the 3 input features for each of the tasks in \figref{fig:single_vs_multitask}.



\subsubsection{Single and Multitask COMPAS Recidivism Prediction}
\label{sec:recidivi_new}

COMPAS is a proprietary score developed to predict recidivism risk, which is used to inform bail, sentencing and parole decisions and has been the subject of scrutiny for racial bias~\citep{propublica, dressel2018accuracy, tan2018distill}.
In 2016, ProPublica released recidivism data~\citep{compas} 
on defendants in Broward County, Florida.

\begin{table*}[t]
\centering
\caption{ROC AUC for multitask and single task NAMs on COMPAS dataset, broken down by gender. Each cell contains the mean AUC $\pm$ one standard deviation obtained via 5-fold cross validation. Higher AUCs are better.}
\label{table:mtlcompas}
\begin{tabular}{| c | c | c | c |} 
 \hline
 Model & COMPAS Women & COMPAS Men & COMPAS Combined  \\ [0.5ex] 
 \hline\hline
 Single Task NAM & 0.716 $\pm$ 0.026 & 0.735 $\pm$ 0.009 & 0.737 $\pm$ 0.010\\ 
 Multitask NAM & 0.723 $\pm$ 0.019 & 0.737 $\pm$ 0.009 & 0.739 $\pm$ 0.010\\
 \hline
\end{tabular}
\vspace{-0.35cm}
\end{table*}

{\bf Single Task Recidivism Prediction:} First, we ask whether this dataset is biased using the transparency of single-task NAMs. Figure~\ref{fig:recidivismmtl} shows the learned single-task NAM which is as accurate as black-box models on this dataset (see Table~\ref{table:classification}). The shape function for race indicates that the learned NAM may be racially biased: Black defendants are predicted to be higher risk for reoffending than white or Asian defendants. This suggests that the recidivism data may be racially-biased. The modularity of NAMs makes it easy to correct this bias by simply removing the contributions learned from the race attribute by zeroing out its mean-centered graph in the learned NAM. Although this would drop the AUC score as we are removing a discriminative feature, it may be a more fair model to use for making bail decisions. It is important to keep potentially offending attributes in the model during training so that the bias can be detected and then removed after training. If the offending variables are eliminated before training, it makes debiasing the model more difficult: if the offending attributes are correlated with other training attributes, the bias is likely to spread to those attributes~\citep{berk2018fairness}. The transparency and modularity of NAMs allows one to detect unanticipated biases in data and makes it easier to correct the bias in the learned model.


{\bf Multitask Recidivism Prediction:} In some settings multitask learning can increase accuracy and intelligibility by learning task-specific shape plots that expose task-specific patterns in the data that would not be learned by single task learning.
We reformulate COMPAS as a multitask problem where recidivism prediction for men and women are treated as separate tasks on a NAM with two outputs.
Indeed, we find that a multitask NAM reveals different relationships between race, charge degree, and recidivism risk for men and women while achieving slightly higher overall accuracy.

The right column of Figure~\ref{fig:recidivismmtl} displays a selection of shape plots learned for a multitask NAM trained on the same data as the single task NAM but with Male and Female as separate output tasks. (The remaining MTL shape plots are similar for the two genders, reinforcing that these are strongly related tasks, but we omit them for brevity.) The race shape plot in the multitask NAM shows a different pattern of racial bias for each gender. The curve for men looks similar to that of the single task NAM (which is expected because men make up 81\% of the data), but the curve for women suggests that recidivism risk is lower for Black women and higher for Caucasian and Hispanic women than for their male counterparts. The multitask shape plots also reveal that charge degree is almost twice as important for women as it is for men. The straightforward extension of NAMs to the multitask setting offers a useful modelling technique not currently available with tree-based GAMs.

\vspace{-0.15cm}
\section{Related Work}
\vspace{-0.15cm}
Generalized Additive Neural Networks~(GANNs)~\citep{potts1999generalized} are somewhat similar to the NAMs we propose here. Like NAMs, GANNs used a restriction in the neural net architecture to force it to learn additive functions of individual features. GANNs, however, predate deep learning and use a single hidden layer with typically only 1-5 hidden units. Furthermore, GANNs did not use backpropagation~\citep{rumelhart1986learning}, required human-in-the-loop evaluation and were not successful in training accurate or scalable GAMs with neural nets. See Section~\ref{sec:gann} for a more detailed overview of GANNs.

In contrast, NAMs in this paper benefit from the advances in deep learning.  They use a large number of hidden units and multiple hidden layers per input feature subnet to allow more complex, more accurate shape functions to be learned. Furthermore, NAMs use novel ExU hidden units to allow subnets to learn the more non-linear functions often required for accurate additive models, and then form an ensemble of these nets to provide uncertainty estimates, further improve accuracy and reduce the high-variance that can result from encouraging the model to learn highly non-linear functions.  

Prior to NAMs, the state-of-the-art in high-accuracy, interpretable generalized additive models~\citep{hastie1990generalized, guisan2002generalized} are the GAM~\citep{lou2012intelligible} and G$\mathrm{A}^2$M~\citep{lou2013intelligible} based on regularized boosted decision trees which were successfully applied to healthcare datasets~\citep{caruana2015intelligible}. We compare the accuracy of NAMs to these models in Section~\ref{sec:experiments}. We note that pairwise interactions, similar to G$\mathrm{A}^2$Ms, can be easily added to NAMs – G$\mathrm{A}^2$Ms use a heuristic to compute the importance of each pairwise interaction by fitting residual from first-order terms and select the k ($\leq 10$) most important interactions. We don’t consider such interactions to keep the paper focused on additive modeling with neural nets. 

\vspace{-0.2cm}
\section{Conclusion and Future Work}
We present Neural Additive Models~(NAMs), which combine the inherent interpretability of GAMs with the expressivity of DNNs, opening the door for other advances in interpretability in deep learning. NAMs are competitive in accuracy to GAMs 
and accurate alternatives to prevalent interpretable models~(\eg shallow trees) 
while being more easily extendable than existing GAMs due to their differentiability and composability. 

A promising direction for future work is improving the expressivity of NAMs by incorporating higher-order feature interactions. While such interactions may result in more expressivity, they might worsen the intelligibility of the learned NAM. Thus, finding a small number of crucial interactions seems important for more expressive yet intelligible NAMs. Another interesting avenue is developing better activation functions or feature representations for easily expressing complex functions using NAMs. For example, fourier features~\citep{tancik2020fourier} have been shown to be highly effective for learning high frequency functions via neural networks and might be useful for training expressive NAMs.

Extending and applying NAMs beyond tabular data to more complex tasks with high-dimensional inputs, such as computer vision and language understanding, is an exciting avenue for future work. While NAMs only use some of the expressivity of DNNs, one can imagine using NAMs in a real-world pipeline where intelligibility is required for decision making from representation~\citep{rudin2021interpretable} (\eg features learned from images, speech etc). Much of the existing interpretability work in deep learning focuses on making learned representations interpretable. Also, NAMs can be used for interpretability across multiple raw features (\eg multimodal inputs) where interpretability within a NAM network can utilize existing interpretability methods in ML – recently CNN-LSTM based extension of NAMs have already been developed for genomics~\citep{srivastava2021interpretable} where the input to each NAM network was a one-hot encoded DNA sequence (passed as an image).  Overall, we believe that NAMs are likely to broaden the use of inherently interpretable models in the deep learning community.

\vspace{-0.2cm}
\section*{Broader Impact}
\vspace{-0.1cm}

Interpretability in AI systems might be desirable or necessary for various reasons -- see \citep{explainable_royalsociety} for an overview; we discuss some of them in the context of NAMs below:
\begin{itemize}[topsep=0pt, partopsep=0pt, leftmargin=13pt, parsep=0pt, itemsep=4pt]
    \item{\bf Safeguarding against bias}: NAMs can check whether training data is used in ways that result in bias or discriminatory outcomes and can be easily corrected for bias to yield possibly more fair models -- \eg Section~\ref{sec:recidivi_new} demonstrates this utility for recidivism risk prediction.
    \item{\bf Improving AI system design}: NAMs allow developers to interrogate why it behaved in a certain way~(\eg tracking system malfunctions), and develop improvements -- Section~\ref{sec:mimic2} shows that NAMs can explain seemingly anomalous results in healthcare as well as uncover problems that might put some kinds of patients at risk and need correction before deploying the system.
    \item {\bf Adhering to regulatory standards or policy requirements}:  Interpretability of NAMs can be important in enforcing legal rights surrounding a system -- \eg credit scores in the United States, have a well-established ``right to explanation''. NAMs can also enable individuals to contest model outputs, \eg challenging an unsuccessful loan application, based on the interpretations provided by NAMs for a specific decision~(Figure~\ref{fig:fico_local}).
    \item {\bf Assessing risk, robustness, and vulnerability}: This can be particularly important if an AI system is deployed in a new environment, where we cannot be sure of its effectiveness -- \eg NAMs for fraud detection~(Section~\ref{sec:fraud}) can be analyzed to understand the risks involved or how it might fail before deploying it to unseen customers.
    \item {\bf Giving users confidence in the system}: Interpretations from NAMs might provide users confidence that it works as intended -- \eg expensive house prices near metropolitan areas such as San Francisco, as predicted by NAMs~(\figref{fig:housing_paper}), is expected for a trustworthy model.
    \item {\bf Data-driven scientific discovery}: NAMs can be applied in natural sciences -- \eg ecology~\citep{guisan2002generalized}, medicine~\citep{hastie1995generalized}, astronomy~\citep{10.1093/mnras/sty3314} \etc. -- to obtain novel scientific insights and discoveries from observational or simulated data~\citep{ai_in_science, roscher2019explainable} while remaining scalable to the ever-increasing data.
\end{itemize}

There are also pitfalls associated with interpretability methods -- NAMs are no exception. Different contexts give rise to different interpretability needs -- \eg public have different expectations of systems used in healthcare \versus recruitment~\citep{project_explain}. Furthermore, AI system designs often need to balance competing demands -- \eg to optimize the accuracy of a system or ensure fairness~(NAMs for making bail decisions with race feature ``removed'' may be less accurate but more fair). In many critical decision-making areas -- \eg healthcare, justice, and public services -- complex processes have developed over time to provide safeguards, audit functions, or other forms of accountability. NAMs may therefore be only the first step in creating trustworthy systems. Those developing NAMs must consider how their use fits in the wider socio-technical context of its deployment



\vspace{-0.1cm}
\section*{Acknowledgments}
We would like to thank Kevin Swersky for reviewing an early draft of the paper. We also thank Sarah Tan for providing us with pre-processed versions of some of the datasets used in the paper. RA would also like to thank Marlos C. Machado and Marc G. Bellemare for helpful discussions.
\bibliography{main}
\bibliographystyle{plainnat}

\section*{Checklist}
\input{checklist}

\appendix
\counterwithin{figure}{section}
\counterwithin{table}{section}

\cleardoublepage

\section{Supplementary Material for Neural Additive Models}


\vspace{-0.23cm}
\subsection{NAMs on MIMIC-II: Mortality Prediction in ICUs}\label{sec:mimic2}
\vspace{-0.07cm}

\input{mimic2}

\subsection{Intelligibility of NAMs on other datasets}
\label{sec:additional_datasets}

\vspace{-0.1cm}
\subsubsection{FICO Score: Understanding Individual Predictions on Credit Scores}
\label{sec:fico}
The FICO score is a widely used proprietary credit score to determine credit worthiness for loans in the United States. The FICO dataset~\citep{fico} is comprised of real-world anonymized credit applications made by customers and their assigned FICO Score, based on their credit report information. 
We visualize the feature contributions of a NAM trained using the FICO dataset~(see Figure~\ref{fig:fico} in appendix) for two applicants~(\tabref{table:fico_person_high}) with low and high scores respectively.  

Figure~\ref{fig:fico_local} shows that the most important features for the high scoring applicant are (1) Average Months on File and (2) Net Fraction Revolving Burden~(\ie percentage of credit limit used) which take the value 235 months and 0\% respectively. This makes sense, as generally, the longer a person's credit history, the better it is for their credit score. Although there is a strong inverse correlation between Net Fraction Revolving Burden and the score, it is positively correlated for small values~($< 10$). This means that making use of some credit increases your credit score, but using too much of it is bad. We are confident in this interpretation because most of the data density is in small values, and each NAM in the ensemble displays a similar shape function~(Figure~\ref{fig:fico}). For the low scoring applicant, the main factors are (1) Total Number of Trades\footnote{Credit trades refer to any agreement between a lending agency and consumers.} and (2) Net Fraction Installment Burden~(Installment balance divided by original loan amount) which take the values 57 and 68\% respectively. This applicant used their credit quite frequently and has a large burden, thus resulting in a low score.

\vspace{-0.25cm}
\subsubsection{Credit Fraud: Financial Fraud Detection~[Classification]}
\label{sec:fraud}
\vspace{-0.05cm}
This is a large dataset~\citep{dal2015adaptive} containing 284,807 transactions made by European credit cardholders where the task is to predict whether a given transaction is fraudulent or not. It is highly unbalanced and contains only 492 frauds~(0.172\% of the entire dataset) of all transactions. \tabref{table:classification} shows that on this dataset, NAMs outperform EBMs and perform comparably to the XGBoost baseline. This shows the benefit of using NAMs instead of tree-based GAMs and suggests that NAMs can provide highly accurate and intelligible models on large datasets. NAMs using ExU units perform much better compared to NAMs with standard DNNs~(AUC~$\approx$~0.974).

\begin{figure*}[t]
      \centering
      \includegraphics[width=\linewidth]{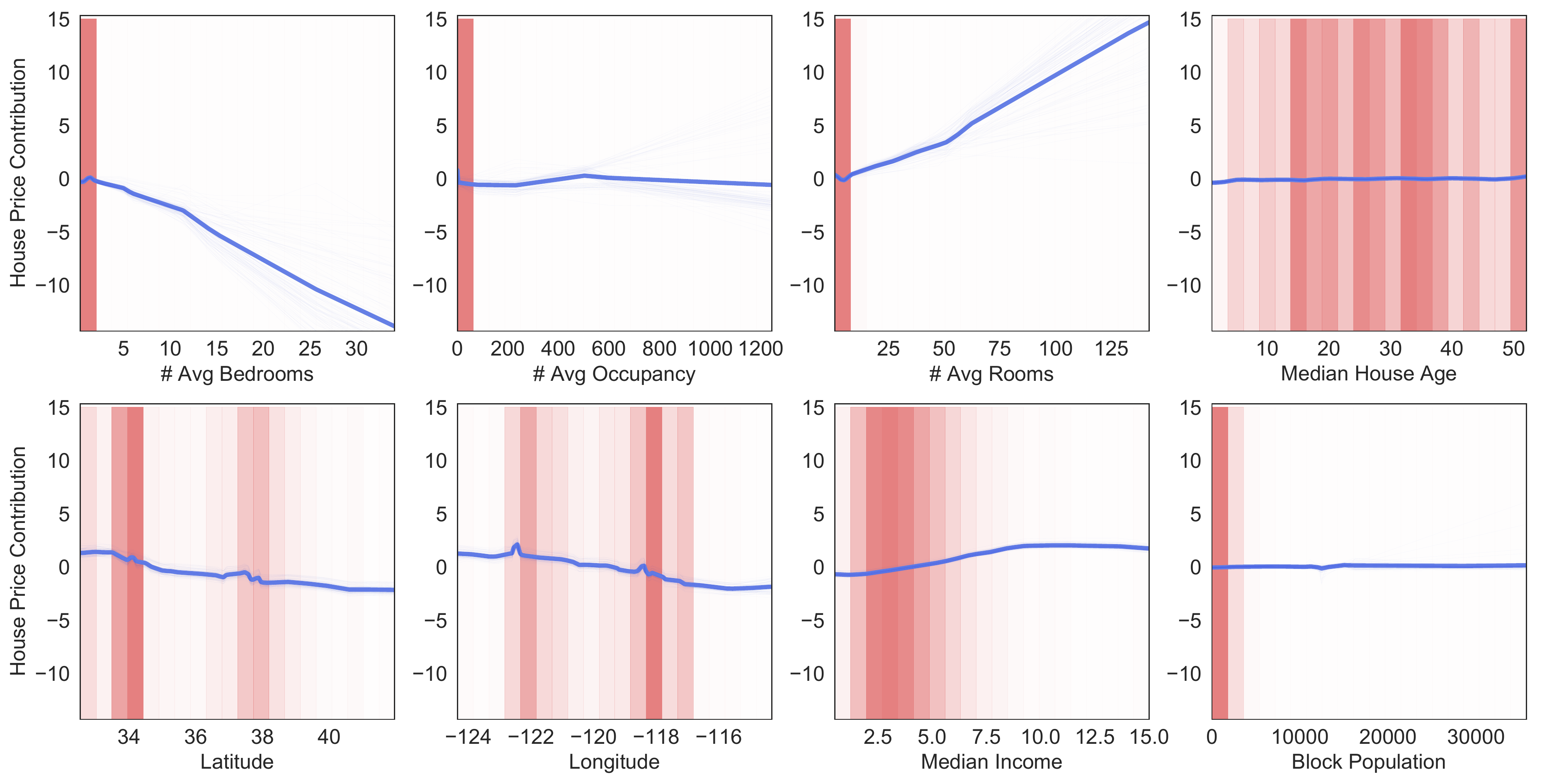}
      \vspace{-0.3cm}
      \caption{{\bf California Housing}. Graphs learned by NAMs trained to predict house prices~(regression) on the California Housing dataset.
      These plots show the individual shape functions learned by an ensemble of hundred NAMs for each input feature as well as the data density. The thin blue lines represents different shape functions from the ensemble to show the agreement of the members of the ensemble. The pink bars represent the normalized data density for each feature. The darker the bar the more data there is with that value.
      }
      \label{fig:housing}
      \vspace{-0.4cm}
\end{figure*}

\vspace{-0.25cm}
\subsubsection{California Housing: Predicting Housing Prices~[Regression]}
\label{sec:housing}
\vspace{-0.05cm}
California Housing dataset~\citep{pace1997sparse} is a canonical machine learning dataset 
where the 
task is to predict the median price of houses~(in million dollars) in each California district. The learned NAM considers the median income as well as the house location~(latitude, longitude) as the most important features (we omit the other six graphs to save space, see Figure~\ref{fig:housing}). As shown by Figure~\ref{fig:housing_paper}, the house prices increase linearly with median income in high data density regions. Furthermore, the graph for longitude shows sharp jumps in price prediction around 122.5$^{\circ}$W and 118.5$^{\circ}$W which roughly correspond to San Francisco and Los Angeles respectively.

\begin{figure}[h]
\centering
\vspace{-0.2cm}
\includegraphics[width=0.6\linewidth]{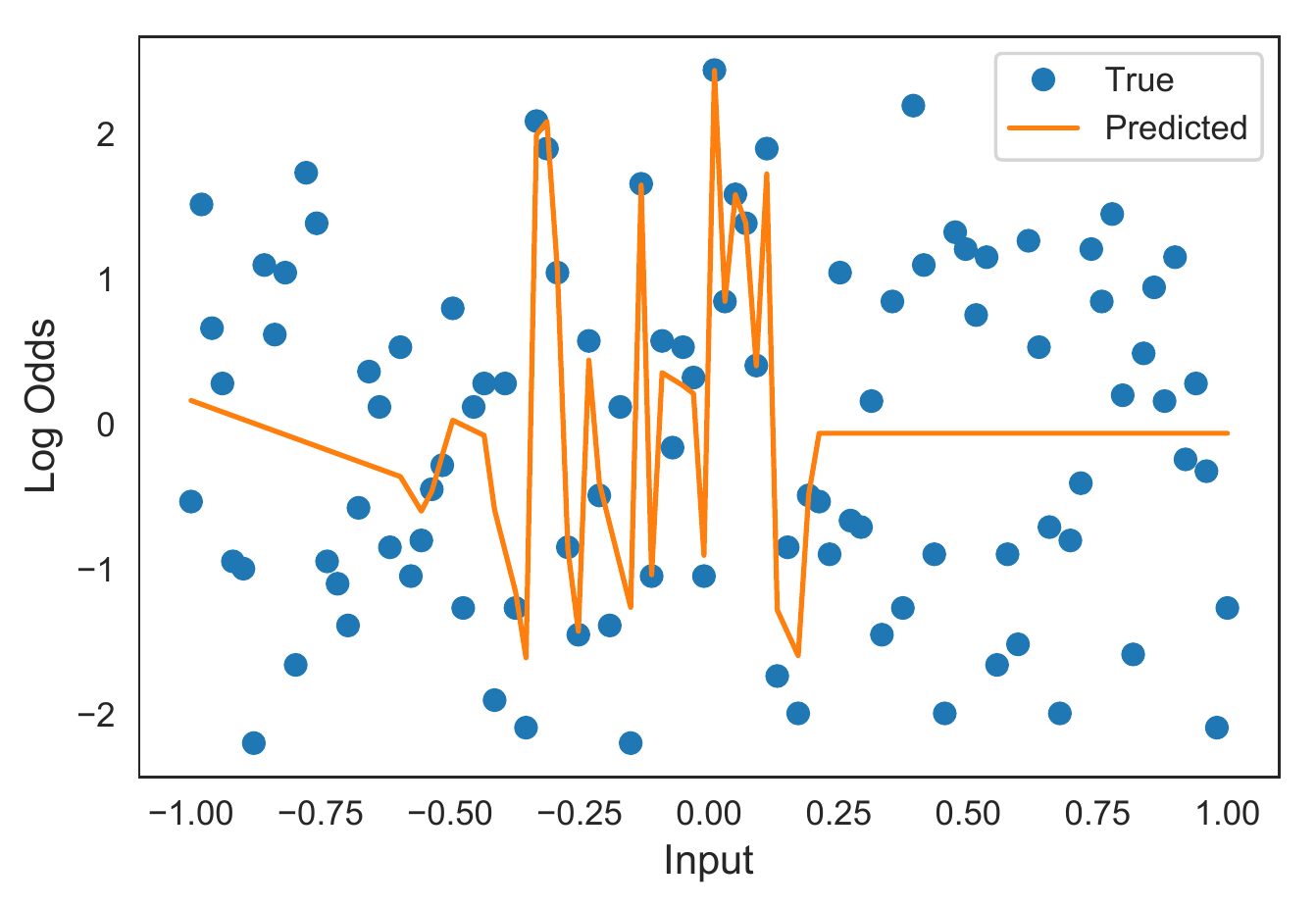}
\caption{{\bf Toy classification}: Deep neural network with 3 hidden layers of size 64, 64 and 32 respectively with ReLU activation and Xavier initialization trained for 10,000 epochs on toy classification 
dataset described in Section~\ref{sec:new_activation}. We use a batch size of 1024 with the Adam optimizer and a learning rate decay of 0.995 every epoch. The learning rate was tuned in [1e-3, 1e-1) and we show the results with the best learning rate.}
\label{fig:dnn_toy}
\end{figure}


\vspace{-0.3cm}
\subsection{Regularization and Training}\label{sec:regularization}
ExU units encourage learning highly jagged curves, however, most realistic shape functions tend to be smooth with large jumps at only a few points. 
To avoid overfitting, we use 
the following regularization techniques:
\begin{itemize}[topsep=0pt, partopsep=0pt, leftmargin=14pt, parsep=0pt, itemsep=4pt]
    \item {\bf Dropout}~\citep{srivastava2014dropout}: It regularizes ExUs in each feature net, allowing them to learn smooth functions while being able to represent jumps~(Figure~\ref{fig:dropout_vs_without}).
    \item {\bf Weight decay}:  This is done by penalizing the L2 norm of weights in each feature net.
    \item {\bf Output Penalty}: We penalize the L2 norm of the prediction of each feature net, so that its contribution stays close to zero unless evident otherwise from the data. 
    \item {\bf Feature Dropout}: We also drop out individual feature networks during training. When there are correlated input features, an additive model can possibly learn multiple explanations by shifting contributions across these features. This term encourages NAMs to spread out those contributions.
\end{itemize}

{\bf Training}. Let $\setD = {\{(\rvx^{(i)}, y^{(i)})\}}_{i=1}^{N}$ be a training dataset of size $N$, where each input $\rvx = (x_1,\ x_2,\ \dots,\ x_K)$ contains $K$ features and $y$ is the target variable.
In this work, we train NAMs using the loss $\calL(\theta)$ given by
\begin{equation}
    \calL(\theta) = \expected_{x, y \sim \setD} \big[l(x, y; \theta) + \lambda_1 \eta(x; \theta)\big] + \lambda_2 \gamma(\theta)
\end{equation}
where $\eta(x; \theta) = \frac{1}{K} \sum_x \sum_{k} ({f^{\theta}_{k}(x_k)})^2$ is the output penalty, $\gamma(\theta)$ is the weight decay and $f^{\theta}_{k}$ is the feature network for the $k^{\mathrm{th}}$ feature.
Each individual network is also regularized using feature dropout and dropout with coefficients $\lambda_3$ and $\lambda_4$ respectively. $l(x, y; \theta)$ is the task dependent loss function. We use the cross-entropy loss for binary classification: 
\begin{equation*}
    l(x, y; \theta) = -y \log (p_{\theta}(x))  -(1 - y) \log (1 - p_{\theta}(x)),
\end{equation*}
where $ p_{\theta}(x) = \sigma\big(\beta^{\theta} + \sum_{k=1}^{K} f^{\theta}_{k}(x_k)\big)$ and mean squared error~(MSE) for regression:
\begin{equation*}
    l(x, y; \theta) = \big(\beta^{\theta} + \sum_{k=1}^{K} f^{\theta}_{k}(x_k) - y \big)^2
\end{equation*}

\subsection{Some practical considerations when using NAMs}

How NAMs performs when the underlying features are additive~(\ie no non-linearities)? We empirically observed that the NAM MLPs do end up approximately recovering the linear functions. That said, the inductive bias of NAMs is toward learning non-linear functions and they might be more expensive than linear models – once a user sees that a NAM learns a linear function for a specific feature, they can try substituting that feature network with a simpler one (or non-linear one) to see if that improves generalization.

Were there instabilities when using ExUs? Surprisingly, we did not observe any instability in training dynamics (across the 4 datasets and synthetic example) and we speculate this is because any small change in weights can lead to significantly peaky function which results in huge loss on training points. Also, we used the Adam optimizer, which adapts the norm of the gradient and prevents them from exploding. Various regularization approaches including weight-regularization and dropout further stabilize the dynamics.

Should we use ExUs vs ReLUs? While a general guidance might be tricky, we hope that ExUs might help certain users to benefit more from the ability of fitting smooth functions but with large jumps at a few point. Devising better activation functions for NAMs is an open research problem.

\subsection{Experimental Details}\label{sec:train}
{\bf Training Details}. The NAM feature networks~($f^{\theta}_{k}$) are trained jointly using the Adam optimizer~\citep{kingma2014adam} with a batch size of 1024 for a maximum of 1000 epochs with early stopping using the validation dataset. The learning rate is annealed by a factor of 0.995 every training epoch. For all the tasks, we tune the learning rate, output penalty coefficient~($\lambda_1$), weight decay coefficient~($\lambda_2$), dropout rate~($\lambda_3$) and feature dropout rate~($\lambda_4$). For computational efficiency, we tune these hyperparameters using Bayesian optimization~\citep{snoek2012practical, golovin2017google} based on cross-validation performance with a single train-validation split for each fold.  We used TESLA P100 GPUs for all experiments involving neural networks while CPU machines for {\it sklearn} or XGBoost baselines.

{\bf Evaluation}. We perform 5-fold cross validation to evaluate the accuracy of the learned models. To measure performance, we use area under the precision-recall curve~(AUC) for binary classification~(as the datasets are unbalanced) and root mean-squared error~(RMSE) for regression. For NAMs and DNNs, one of the 5 folds~(20\% data) is used as a held-out test set while the remaining 4 folds are used for training~(70\% data) and validation~(10\% data). The training and validation splits are randomly subsampled from the 4 folds and this process is repeated 20 times. For each run, the validation set is used for early stopping. For each fold, we ensemble the NAMs and DNNs trained on the 20 and EBMs on 100 train-validation splits respectively to make the prediction on the held-out test set.

\subsection{Hyperparameters}
We use a batch size of 1024 with the Adam optimizer and a learning rate decay of 0.995 every epoch in our experiments for NAMs and DNNs.

{\bf Linear Models/ Decision Trees}. We use the \textit{sklearn} implementation~\citep{pedregosa2011scikit}, and tune the hyperparameters with grid search.

{\bf EBMs}. We use the open-source implementation~\citep{nori2019interpretml} with the parameters specified by prior work~\citep{caruana2015intelligible} for a fair comparison.

{\bf NAMs}. We tune the dropout coefficient~($\lambda_3$) in the discrete set \{0, 0.05, 0.1, 0.2, 0.3, 0.4, 0.5, 0.6, 0.7, 0.8, 0.9\}, weight decay coefficient~($\lambda_2$) in the continuous interval $[0.000001, 0.0001)$, learning rate in the interval $[0.001, 0.1)$, feature dropout coefficient~($\lambda_4$) in the discrete set \{0, 0.05, 0.1, 0.2\} and output penalty coefficient~($\lambda_1$) in the interval $[0.001, 0.1)$. Note that the weight decay is implemented as the average weight decay over the individual feature networks in NAMs. Refer to \tabref{table:clasification_hyperparams} and \tabref{table:regression_hyperparams} for hyperparameters found on regression and classification datasets.
 
{\bf DNNs}. We train DNNs with 10 hidden layers containing 100 units each with ReLU activation using the Adam optimizer. This architecture choice ensures that this network had the capacity to achieve perfect training accuracy on datasets used in our experiments. We use weight decay and dropout to prevent overfitting and tune hyperparameters using a similar protocol as NAMs. We tune the dropout coefficient~($\lambda_3$) in \{0, 0.05, 0.1, 0.2, 0.3, 0.4, 0.5\}, weight decay coefficient~($\lambda_2$) in the continuous interval $[0.0000001, 0.1)$ and learning rate in the interval $[0.001, 0.1)$.

\begin{table}[h]
\centering
\caption{Optimal hyperparameters found for NAMs on regression datasets. ``Hidden units'' shows the number of hidden layers as well as the number of neurons used in each layer for each feature network.}
\label{table:regression_hyperparams}
\begin{tabular}{| c | c | c | c |} 
 \hline
 Hyperparameter & FICO & Housing \\ [0.5ex] 
 \hline\hline
 Learning Rate & 0.0161 & 0.00674 \\
 \hline
 Output Penalty~($\lambda_1$) & 0.0205 & 0.001 \\
 \hline
 Weight Decay~($\lambda_2$) & 1.07 x $10^{-5}$ & $10^{-6}$ \\
 \hline
 Dropout & 0.0 & 0.0 \\
 \hline
 Feature Dropout  & 0.0 & 0.0 \\ 
 \hline
 Num units & 64, 64, 32 & 64, 64, 32 \\
  \hline
 Activation & ReLU & ReLU \\
  \hline
 Hidden unit & Standard & Standard \\
 \hline
\end{tabular}

\end{table}

\begin{table}[h]
\centering
\caption{Optimal hyperparameters found for NAMs on classification datasets. ``Hidden units'' shows the number of hidden layers as well as the number of units used in each hidden layer for each feature network.}
\vspace{0.2cm}
\label{table:clasification_hyperparams}
\begin{tabular}{| c | c | c | c |} 
 \hline
 Hyperparameter & COMPAS & MIMIC-II & Credit Fraud  \\ [0.5ex] 
 \hline\hline
 Learning Rate & 0.02082  & 0.005 & 0.0157 \\
 \hline
 Output Penalty & 0.2078 & 0.3 & 0.0 \\
 \hline
 Weight Decay & 0.0 & 9.6 x $10^{-5}$ & 4.95 x $10^{-6}$\\
 \hline
 Dropout & 0.1 & 0.2 & 0.8 \\
 \hline
 Feature Dropout & 0.05 & 0.0 & 0.0\\ 
 \hline
 Num units & 64, 64, 32 & 1024 & 1024 \\
 \hline
 Activation & ReLU & ReLU-1 & ReLU-1 \\
  \hline
 Hidden unit & Standard & ExU & ExU \\
 \hline
\end{tabular}
\end{table}

\subsection{GANNs}
\label{sec:gann}
GANNs begin with subnets containing a single hidden unit for each input feature, and use a human-in-the-loop process to add (or subtract) hidden units to the architecture based on human evaluation of plotted partial residuals.  This means that the training procedure cannot be automated. In practice, the laborious manual effort required to evaluate all of the partial residual plots to decide what to do next, and then retrain the model after adding or subtracting hidden units from the architecture meant that GANN nets remained very small and simple --- typically only one hidden unit per feature.





\begin{table*}[h]
\centering
\caption{Feature attributes for the two individuals shown in Figure~\ref{fig:fico_local}}
\label{table:fico_person_high}
\begin{tabular}{| c | c | c |} 
 \hline
 Feature & High Score Applicant & Low Score applicant \\ [0.5ex] 
 \hline\hline
Months Since Oldest Trade Open & 417.0 & 174.0 \\
Months Since Most Recent Trade & 25.0 & 1.0 \\
Average Months in File & 235.0 & 66.0 \\
\# Satisfactory Trades & 9.0 & 44.0 \\
\# Trades 60+ Ever & 0.0 & 11.0 \\
\# Trades 90+ Ever & 0.0 & 8.0 \\
\% Trades Never Delinquent & 100.0 & 70.0 \\
Months Since Most
 Recent Delinquency & 0.0 & 3.0 \\
Max Delq/Public Records
 Last Year & 6.0 & 0.0 \\
Max Delinquency Ever & 6.0 & 0.0 \\
\# Total Trades & 9.0 & 57.0 \\
\# Trades Open in Last 12 Months & 0.0 & 5.0 \\
\% Installment Trades & 22.0 & 66.0 \\
Months Since Most Recent
Inquiry excluding 7 days & 0.0 & 0.0 \\
\# Inquiries in Last 6 Months & 1.0 & 7.0 \\
\# Inquiries in Last 6 Months 
 excluding 7 days & 0.0 & 6.0 \\
Net Fraction Revolving Burden & 0.0 & 23.0 \\
Net Fraction Installment Burden & 0.0 & 68.0 \\
\# Revolving Trades with Balance & 1.0 & 2.0 \\
Number Installment Trades
 with Balance & 0.0 & 5.0 \\
\# Bank/Natl Trades with
 high utilization ratio & 0.0 & 0.0 \\
\% Trades with Balance & 40.0 & 64.0 \\
Delinquent & 0.0 & 1.0 \\
Inquiry & 1.0 & 1.0 \\
 \hline
\end{tabular}
\end{table*}

\begin{table}[t]
\centering
\caption{{\bf FICO Score}. Meaning of the different attributes of the feature ``Max Delq/Public Records Last Year''.}
\vspace{0.1cm}
\label{table:MaxDelq2PublicRecLast12M}
\begin{tabular}{| c | c |} 
 \hline
 Value & Meaning  \\ [0.5ex] 
\hline\hline
0 & Derogatory comment \\
1 & 120+ days delinquent \\
2 & 90 days delinquent \\
3 & 60 days delinquent \\
4 & 30 days delinquent \\
5,6 &   Unknown delinquent \\
7 & Current and never delinquent \\
8,9 & All other \\
 \hline
\end{tabular}
\end{table}

\begin{figure*}[t]
      \centering
      \includegraphics[width=\linewidth]{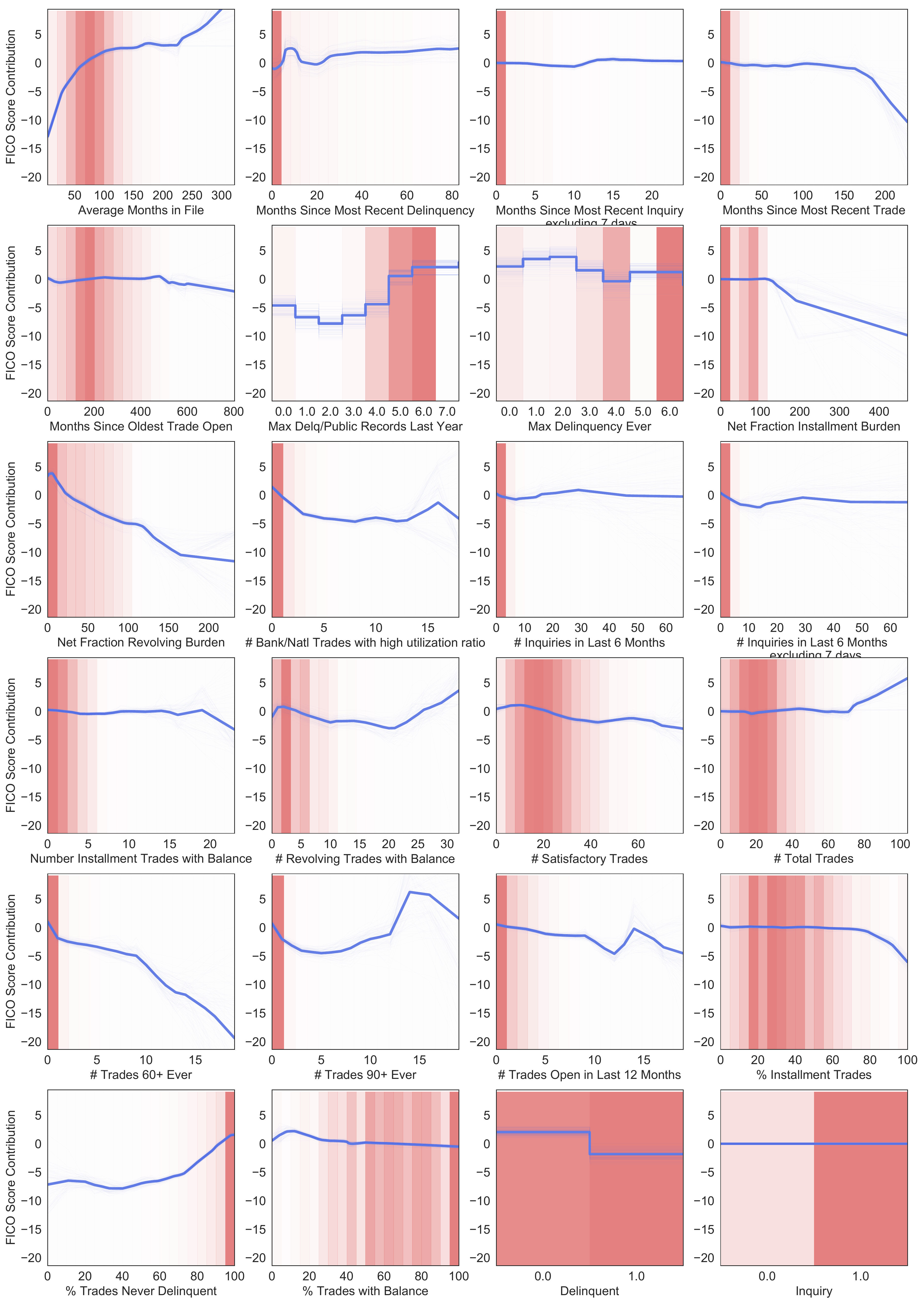}
      \vspace{-0.1cm}
      \caption{{\bf FICO Score Prediction}. Graphs learned by NAMs trained to predict FICO scores~(regression) based on their credit report information.
      These graphs can be interpreted easily, \eg the second last graph in the bottom row shows that being delinquent on your payments decreases your credit score.
      }
      \label{fig:fico}
\end{figure*}

\begin{figure*}[t]
      \centering
      \includegraphics[width=\linewidth]{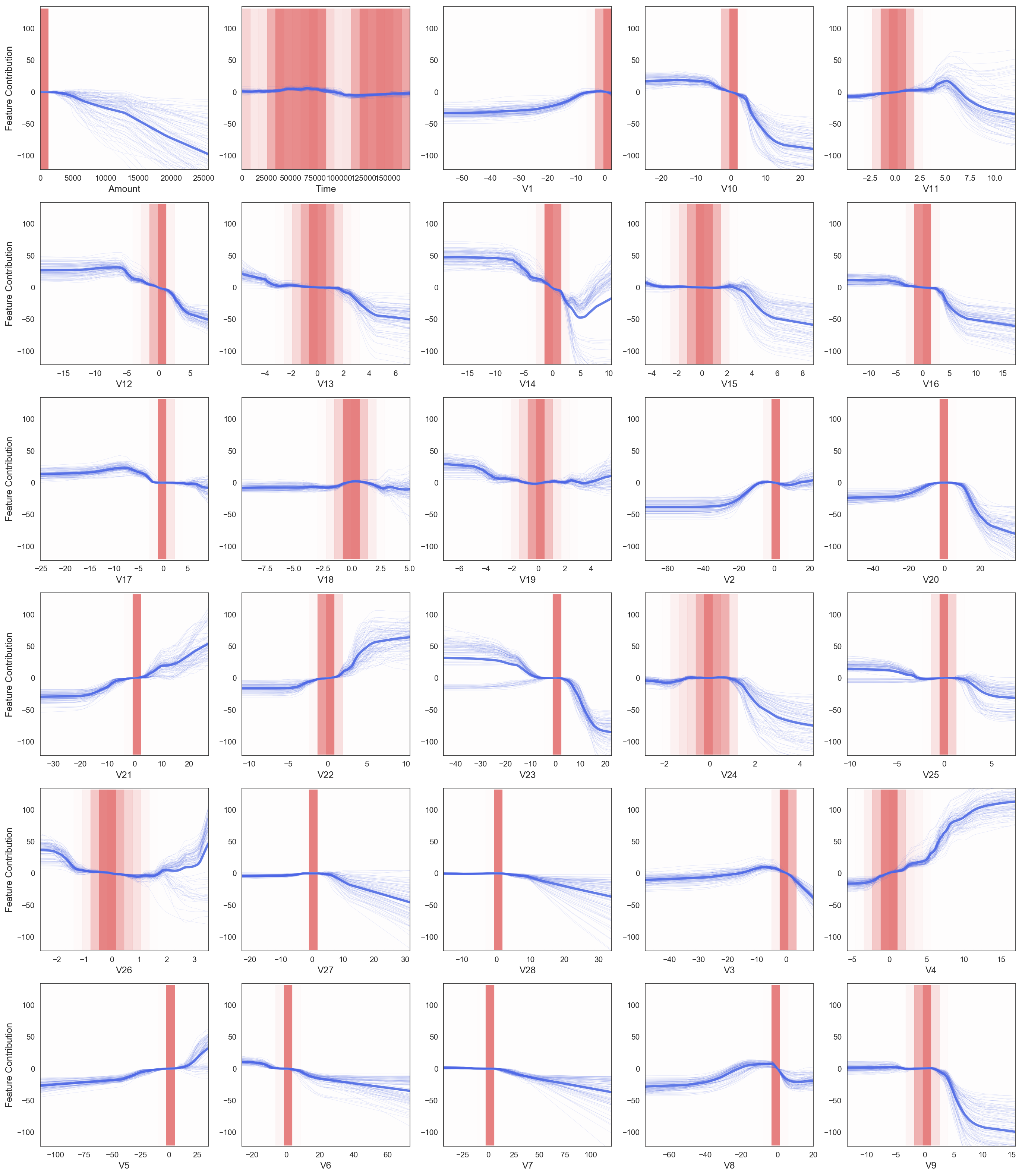}
      \vspace{-0.2cm}
      \caption{\textbf{Credit Fraud Detection}: Graphs learned by NAMs with ExU units on this large classification dataset. The task is to predict credit fraud where the class variable takes value 1 in case of fraud and 0 otherwise using a large dataset of credit card transactions. The dataset only contains only numerical input variables which are the result of a PCA transformation except the features 'Time' and 'Amount'. Unfortunately, due to confidentiality issues, the original features are not provided in the dataset.}
      \label{fig:credit}
\end{figure*}

\vspace{-0.15cm}
\subsection{Multitask Learning}\label{sec:multitask_nams_appendix}

\vspace{-0.15cm}\label{sec:synthetic_generators}
\subsubsection{Synthetic Data Generation}

We used the following generator functions to produce our synthetic dataset:

\begin{equation} \label{eq:generatorfns}
\begin{split}
f(x_0) & = \frac{1}{3} \log{100x_0 + 101} \\ 
g(x_1) & = -\frac{4}{3} e^{-4|x_1|} \\ 
h(x_2) & = \sin{(10x_2)} \\ 
i(x_2) & = \cos{(15x_2)} \\
\end{split}
\end{equation}

Noise sampled from $N(0, \frac{5}{6})$ was added to the target for each task. We will provide our generation code for others who are interested in using this dataset.

\subsubsection{Gains from multi-task NAMs}
We also ran control experiments where we provide multiple subnets for each feature and each task for single-task learning (STL), and this does sometimes improve test accuracy marginally for STL. However, this still performed worse than multi-task NAMs as they are able to make use of samples across multiple tasks to learn a common function but the single-task NAM can’t share samples and don’t have access to enough data for learning individual shape functions.

\vspace{-0.15cm}
\subsubsection{Shape Plots for All Synthetic Features}

\begin{figure*}[t]
      \centering
      \includegraphics[width=\linewidth]{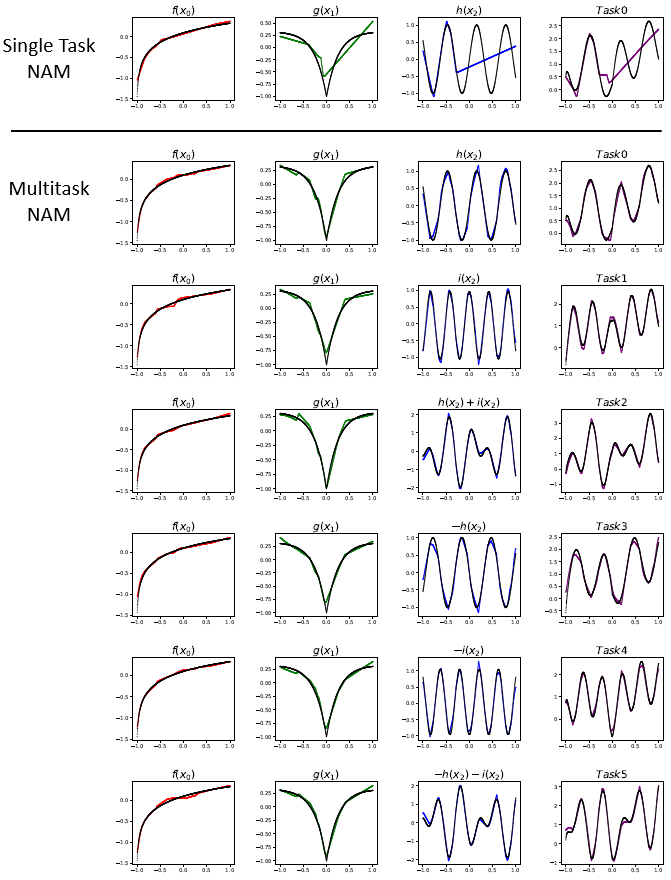}
      \vspace{-0.2cm}
      \caption{\textbf{Single and Multitask NAMs trained on synthetic data}: Shape plots for all synthetic features for a typical (median) run of single and multitask NAMs. The colored lines represent learned shape functions for each feature and the black line represents the generator function.}
      \label{fig:single_vs_multitask}
\end{figure*}

As shown in \figref{fig:single_vs_multitask}, we include here shape plots for all features in the synthetic data for both single and multitask NAMs. The MTL results represent a single model trained on all 6 tasks. In each case, it models the shape functions for every feature and the target with high accuracy. By contrast, the STL for $Task0$, struggles to fit the data for $x_2$ and achieves low accuracy on the target in this regime of noise and training set size. 

\end{document}

%% file: defs.tex
\usepackage{dsfont}
\usepackage{amsthm}
\usepackage{amsmath}
\usepackage{amssymb}
\usepackage{color}
\usepackage{xspace}
\usepackage{pgfplots, pgfplotstable}
\usepackage{xcolor,colortbl}

\def\expected{\mathbb{E}}

\newcommand{\R}[1]{R}
\newcommand{\G}[1]{G}

\newcommand{\Ad}[1]{A}

\newcommand{\CC}[1]{C}

\def\setD{\mathcal{D}}

\def\calL{\mathcal{L}}

\def\@onedot{\ifx\@let@token.\else.\null\fi\xspace}
\def\etc{{\em etc}}
\def\eg{{\em e.g.,}\xspace}
\def\ie{{\em i.e.,}\xspace}

\def\versus{{\em vs.}\xspace}

\newcommand{\figref}[1]{Figure~\ref{#1}}

\newcommand{\tabref}[1]{Table~\ref{#1}}

\usepackage{amsmath,amsfonts,bm}

\def\Secref#1{Section~\ref{#1}}

\def\1{\bm{1}}

\def\rvx{{\mathbf{x}}}

\DeclareMathAlphabet{\mathsfit}{\encodingdefault}{\sfdefault}{m}{sl}
\SetMathAlphabet{\mathsfit}{bold}{\encodingdefault}{\sfdefault}{bx}{n}



%% file: checklist.tex

\begin{enumerate}

\item For all authors...
\begin{enumerate}
  \item Do the main claims made in the abstract and introduction accurately reflect the paper's contributions and scope?
    \answerYes{}
  \item Did you describe the limitations of your work?
    \answerYes{}
  \item Did you discuss any potential negative societal impacts of your work?
    \answerYes{}
  \item Have you read the ethics review guidelines and ensured that your paper conforms to them?
    \answerYes{}
\end{enumerate}

\item If you are including theoretical results...
\begin{enumerate}
  \item Did you state the full set of assumptions of all theoretical results?
    \answerNA{}
	\item Did you include complete proofs of all theoretical results?
    \answerNA{}
\end{enumerate}

\item If you ran experiments...
\begin{enumerate}
  \item Did you include the code, data, and instructions needed to reproduce the main experimental results (either in the supplemental material or as a URL)?
    \answerYes{}
  \item Did you specify all the training details (e.g., data splits, hyperparameters, how they were chosen)?
    \answerYes{}
	\item Did you report error bars (e.g., with respect to the random seed after running experiments multiple times)?
    \answerYes{}
	\item Did you include the total amount of compute and the type of resources used (e.g., type of GPUs, internal cluster, or cloud provider)?
    \answerYes{}
\end{enumerate}

\item If you are using existing assets (e.g., code, data, models) or curating/releasing new assets...
\begin{enumerate}
  \item If your work uses existing assets, did you cite the creators?
    \answerYes{}
  \item Did you mention the license of the assets?
    \answerYes{}
  \item Did you include any new assets either in the supplemental material or as a URL? \answerNA{}
  \item Did you discuss whether and how consent was obtained from people whose data you're using/curating?
    \answerYes{}
  \item Did you discuss whether the data you are using/curating contains personally identifiable information or offensive content?
    \answerYes{}
\end{enumerate}

\item If you used crowdsourcing or conducted research with human subjects...
\begin{enumerate}
  \item Did you include the full text of instructions given to participants and screenshots, if applicable?
    \answerNA{}
  \item Did you describe any potential participant risks, with links to Institutional Review Board (IRB) approvals, if applicable?
  \answerNA{}
  \item Did you include the estimated hourly wage paid to participants and the total amount spent on participant compensation?
  \answerNA{}
\end{enumerate}

\end{enumerate}

%% file: mimic2.tex
\begin{figure*}[h]
  \centering
  \includegraphics[width=\linewidth]{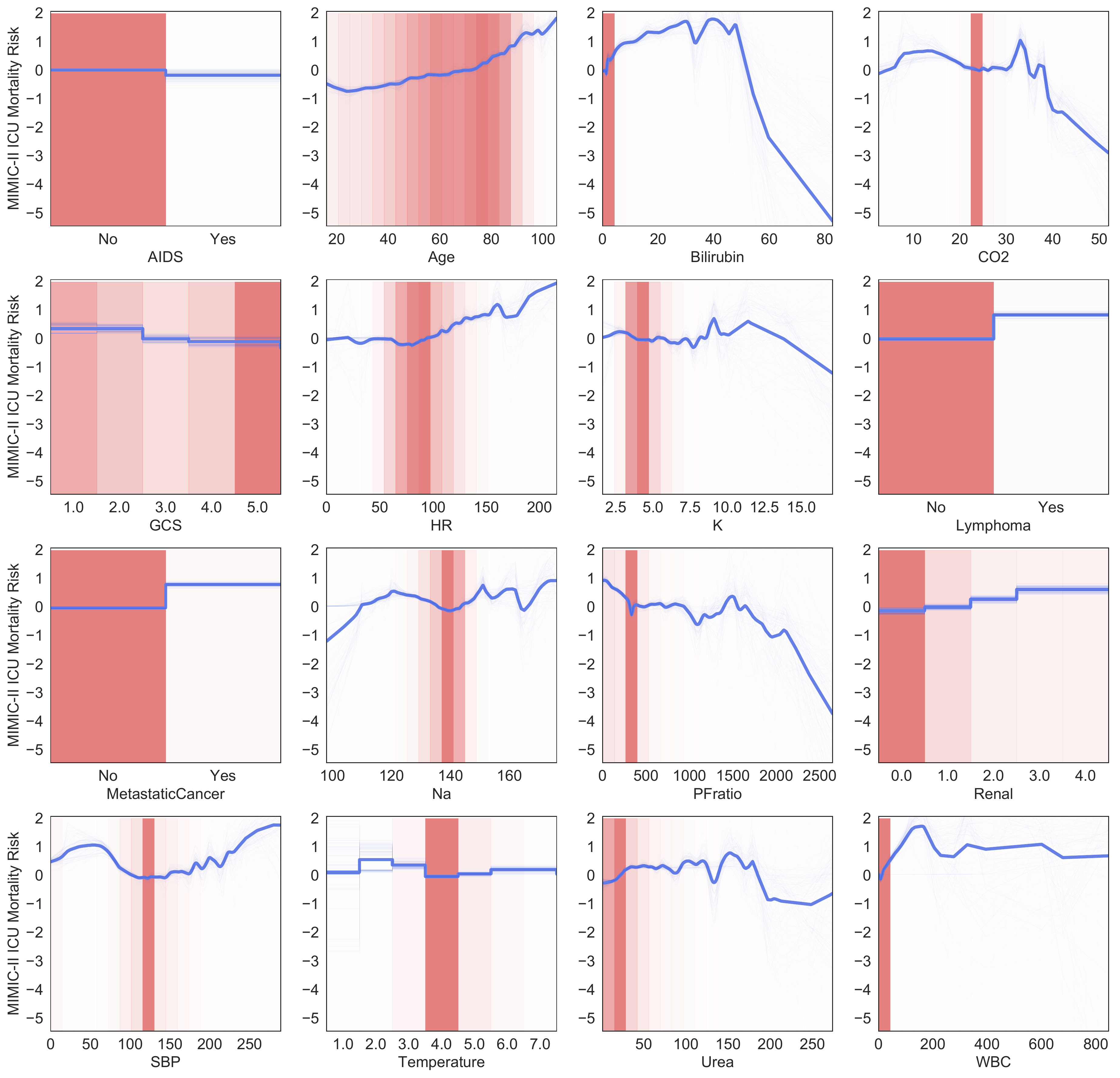}
  \vspace{-0.3cm}
  \caption{{\bf MIMIC-II ICU Mortality}. NAM shape functions learned on the MIMIC-II dataset to predict mortality risk using medical features~(shown on the $x$-axis) collected during the stay in the ICU. Low values on the $y$-axis indicates a low risk of mortality.}
  \label{fig:mimic2}
  \vspace{-0.1cm}
\end{figure*}

Figure~\ref{fig:mimic2} shows 16 of the shape functions learned by the NAM for the MIMIC-II dataset~\citep{saeed2011multiparameter} to predict mortality in intensive care unit~(ICUs). (The $17^{th}$ graph for Admission Type is flat and we omit it to save space.) The plot for HIV/AIDS shows that patients with AIDS have less risk of ICU mortality. While this might seem counter-intuitive, we confirmed with doctors that this is probably correct: among the various reasons why one might be admitted to the ICU, AIDS is a relatively treatable illness and is one of the less risky reasons for ICU admission. In other words, being admitted to the ICU for AIDS suggests that the patient was not admitted for another riskier condition, and thus the model correctly predicts that the patient is at less risk than other non-AIDS patients.  

The shape plot for Age shows, as expected, that mortality risk tends to increase with age, with the most rapid rise in risk happening above age 80.  There is detail in the graph that is interesting and warrants further study such as the small increase in risk at ages 18 and 19, and the bump in risk at age 90 --- jumps in risk that happen at round numbers are often due to social effects. 

The shape plot for Bilirubin (a by product of the breakdown of red blood cells) shows that risk is low for normal levels below 2-3, and rises significantly for levels above 15-20, until risk drops again above 50. There is also a surprising drop in risk near 35 that requires further investigation.  We believe the drop in risk above 50 is because patients above 50 begin to receive dialysis and other critical care and these treatments are very effective. The drop in risk that occurs for Urea above 175 is also likely due to dialysis.

The plot for the Glasgow Coma Index~(GCS) is monotone decreasing as would be expected: higher GCS indicates less severe coma. Note that NAMs are not biased to learn monotone functions such as this and the shape of the plot is driven by the data.  The NAM also learns a monotone increasing shape plot for risk as a function of renal function.  This, too, is as expected: 0.0 codes for normal renal function and 4.0 indicates severe renal failure. 

The NAM has learned that risk is least for normal heart rate~(HR) in the range 60-80, and that risk rises as heart rate climbs above 100. Also, as expected, both Lymphoma and Metastatic Cancer increase mortality risk.    The CO2 graph shows low risk for the normal range 22-24.  
There is an interesting drop in risk at CO2 equal to 37 (the dip between the peaks at 33 and 39) that warrants further investigation.

    
    

The shape plot for PFratio (a measure of the effectiveness of converting O2 in air to O2 in blood) shows a drop at PFratio = 332 which upon further inspection is due to missing values in PFratio being imputed with the mean: because most patients do not have their PFratio measured, the highest density of patients are actually missing their PFratio which was then imputed with the mean value of 332. 
One way to detect that imputed missing values are responsible for a dip (or rise) in a shape plot is when risk at the mean value of the attribute suddenly drops (or rises) to a risk level similar to what the model learns for patients who are considered normal/healthy in this dimension: the jump happens at the mean when imputation is done with the mean value, and the level jumps towards the risk of normal healthy patients because often the variable was not measured because the patients were considered normal (for this attribute), a medical assessment which is often correct.  

Normal temperature is coded as 4.0 on the temperature plot, and risk rises for abnormal temperature above and below this value.  It's not clear if the non-monotone risk for hypothermic patients with temperatures 1 or 2 is due to variance, an unknown problem with the data, or an unexplained but real effect in the training signals and warrants further investigation.  Similarly, the shape plot for Systolic Blood Pressure (SBP) shows lowest risk for normal SBP near 120, with risk rising for abnormally elevated or depressed SBP. The jumps in risk that happen at 175, 200, and 225 are probably due to treatments that doctors start to apply at these levels: the risk rises to left of these thresholds as SBP rises to more dangerous levels but before the treatment threshold is reached, and then drops a little to the right of these thresholds when most patients above the treatment threshold are receiving a more aggressive treatment that is effective at lowering their risk.

{\bf Discussion}. In summary, most of what the NAM has learned appears to be consistent with medical knowledge, though a few details on some of the graphs (e.g., the increase in risk for young patients, and the drop in risk for patients with Bilirubin near 35) require further investigation.  NAMs are attractive models because they often are very accurate, while remaining interpretable, and if a detail in some graph is found to be incorrect, the model can be edited by re-drawing the graph. However,  NAMs~(like all GAMs), are not causal models.  Although the shape plots can be informative, and can help uncover problems with the data that might need correction before deploying the models, the plots do not tell us {\em why} the model learned what it did, or what the impact of intervention (\eg actively lowering a patient's fever or blood pressure) would be. The shape plots do, however, tell us {\em exactly} how the model makes its predictions.

%% file: main.bbl
\begin{thebibliography}{46}
\providecommand{\natexlab}[1]{#1}
\providecommand{\url}[1]{\texttt{#1}}
\expandafter\ifx\csname urlstyle\endcsname\relax
  \providecommand{\doi}[1]{doi: #1}\else
  \providecommand{\doi}{doi: \begingroup \urlstyle{rm}\Url}\fi

\bibitem[Angwin et~al.(2016)Angwin, Larson, Kirchner, and Mattu]{propublica}
Julia Angwin, Jeff Larson, Lauren Kirchner, and Surya Mattu.
\newblock {Machine Bias: There’s software used across the country to predict
  future criminals. And it's biased against blacks}, 2016.
\newblock URL
  \url{https://www.propublica.org/article/machine-bias-risk-assessments-in-criminal-sentencing}.
\newblock [Accessed February 1, 2020].

\bibitem[Arpit et~al.(2017)Arpit, Jastrz{k{e}}bski, Ballas, Krueger, Bengio,
  Kanwal, Maharaj, Fischer, Courville, Bengio, et~al.]{arpit2017closer}
Devansh Arpit, Stanis{l}aw Jastrz{k{e}}bski, Nicolas Ballas, David Krueger,
  Emmanuel Bengio, Maxinder~S Kanwal, Tegan Maharaj, Asja Fischer, Aaron
  Courville, Yoshua Bengio, et~al.
\newblock A closer look at memorization in deep networks.
\newblock \emph{ICML}, 2017.

\bibitem[Berk et~al.(2018)Berk, Heidari, Jabbari, Kearns, and
  Roth]{berk2018fairness}
Richard Berk, Hoda Heidari, Shahin Jabbari, Michael Kearns, and Aaron Roth.
\newblock Fairness in criminal justice risk assessments: The state of the art.
\newblock \emph{Sociological Methods \& Research}, 2018.

\bibitem[Caruana(1997)]{caruana1997mtl}
Rich Caruana.
\newblock Multitask learning.
\newblock \emph{Machine Learning}, 1997.

\bibitem[Caruana et~al.(2015)Caruana, Lou, Gehrke, Koch, Sturm, and
  Elhadad]{caruana2015intelligible}
Rich Caruana, Yin Lou, Johannes Gehrke, Paul Koch, Marc Sturm, and Noemie
  Elhadad.
\newblock Intelligible models for healthcare: Predicting pneumonia risk and
  hospital 30-day readmission.
\newblock \emph{SIGKDD}, 2015.

\bibitem[Chen and Guestrin(2016)]{chen2016xgboost}
Tianqi Chen and Carlos Guestrin.
\newblock {XGBoost}: A scalable tree boosting system.
\newblock \emph{SIGKDD}, 2016.

\bibitem[Dal~Pozzolo(2015)]{dal2015adaptive}
Andrea Dal~Pozzolo.
\newblock Adaptive machine learning for credit card fraud detection.
\newblock \emph{PhD Thesis, Department of Computer Science, Universit{\'e}
  Libre de Bruxelles}, 2015.

\bibitem[Dressel and Farid(2018)]{dressel2018accuracy}
Julia Dressel and Hany Farid.
\newblock The accuracy, fairness, and limits of predicting recidivism.
\newblock \emph{Science advances}, 2018.

\bibitem[{FICO}(2018)]{fico}
{FICO}.
\newblock {FICO Explainable Machine Learning Challenge}.
\newblock
  \url{https://community.fico.com/s/explainable-machine-learning-challenge},
  2018.

\bibitem[Glorot and Bengio(2010)]{glorot2010understanding}
Xavier Glorot and Yoshua Bengio.
\newblock Understanding the difficulty of training deep feedforward neural
  networks.
\newblock \emph{AISTATS}, 2010.

\bibitem[Golovin et~al.(2017)Golovin, Solnik, Moitra, Kochanski, Karro, and
  Sculley]{golovin2017google}
Daniel Golovin, Benjamin Solnik, Subhodeep Moitra, Greg Kochanski, John Karro,
  and D~Sculley.
\newblock Google vizier: A service for black-box optimization.
\newblock \emph{SIGKDD}, 2017.

\bibitem[Guisan et~al.(2002)Guisan, Edwards~Jr, and
  Hastie]{guisan2002generalized}
Antoine Guisan, Thomas~C Edwards~Jr, and Trevor Hastie.
\newblock Generalized linear and generalized additive models in studies of
  species distributions: setting the scene.
\newblock \emph{Ecological modelling}, 2002.

\bibitem[Hastie and Tibshirani(1995)]{hastie1995generalized}
T~Hastie and R~Tibshirani.
\newblock Generalized additive models for medical research.
\newblock \emph{Statistical methods in medical research}, 1995.

\bibitem[Hastie and Tibshirani(1990)]{hastie1990generalized}
Trevor Hastie and Robert Tibshirani.
\newblock \emph{Generalized Additive Models}.
\newblock Chapman and Hall/CRC, 1990.

\bibitem[Hattab et~al.(2018)Hattab, de Souza, Ciardi, Paardekooper, Khochfar,
  and Dalla Vecchia]{10.1093/mnras/sty3314}
M~W Hattab, R~S de Souza, B~Ciardi, J-P Paardekooper, S~Khochfar, and
  C~Dalla Vecchia.
\newblock {A case study of hurdle and generalized additive models in astronomy:
  the escape of ionizing radiation}.
\newblock \emph{Monthly Notices of the Royal Astronomical Society}, 2018.

\bibitem[He et~al.(2015)He, Zhang, Ren, and Sun]{he2015delving}
Kaiming He, Xiangyu Zhang, Shaoqing Ren, and Jian Sun.
\newblock Delving deep into rectifiers: Surpassing human-level performance on
  imagenet classification.
\newblock \emph{CVPR}, 2015.

\bibitem[He et~al.(2016)He, Zhang, Ren, and Sun]{he2016deep}
Kaiming He, Xiangyu Zhang, Shaoqing Ren, and Jian Sun.
\newblock Deep residual learning for image recognition.
\newblock \emph{CVPR}, 2016.

\bibitem[Hornik et~al.(1989)Hornik, Stinchcombe, White,
  et~al.]{hornik1989multilayer}
Kurt Hornik, Maxwell Stinchcombe, Halbert White, et~al.
\newblock Multilayer feedforward networks are universal approximators.
\newblock \emph{Neural Networks}, 1989.

\bibitem[{ICO}(2019)]{project_explain}
{ICO}.
\newblock {Project explain: Interim report}.
\newblock 2019.
\newblock \url{https://ico.org.uk/media/2615039/project-explain-20190603.pdf}.

\bibitem[Kingma and Ba(2014)]{kingma2014adam}
Diederik~P Kingma and Jimmy Ba.
\newblock Adam: A method for stochastic optimization.
\newblock \emph{arXiv preprint arXiv:1412.6980}, 2014.

\bibitem[Krizhevsky(2010)]{krizhevsky2010convolutional}
Alex Krizhevsky.
\newblock Convolutional deep belief networks on cifar-10.
\newblock 2010.

\bibitem[Lee et~al.(2021)Lee, Samad, Hofer, Cannesson, and
  Baldi]{lee2021development}
Christine~K Lee, Muntaha Samad, Ira Hofer, Maxime Cannesson, and Pierre Baldi.
\newblock Development and validation of an interpretable neural network for
  prediction of postoperative in-hospital mortality.
\newblock \emph{NPJ digital medicine}, 2021.

\bibitem[Lou et~al.(2012)Lou, Caruana, and Gehrke]{lou2012intelligible}
Yin Lou, Rich Caruana, and Johannes Gehrke.
\newblock Intelligible models for classification and regression.
\newblock \emph{SIGKDD}, 2012.

\bibitem[Lou et~al.(2013)Lou, Caruana, Gehrke, and Hooker]{lou2013intelligible}
Yin Lou, Rich Caruana, Johannes Gehrke, and Giles Hooker.
\newblock Accurate intelligible models with pairwise interactions.
\newblock \emph{SIGKDD}, 2013.

\bibitem[Nair and Hinton(2010)]{nair2010rectified}
Vinod Nair and Geoffrey~E Hinton.
\newblock Rectified linear units improve restricted boltzmann machines.
\newblock \emph{ICML}, 2010.

\bibitem[Nori et~al.(2019)Nori, Jenkins, Koch, and
  Caruana]{nori2019interpretml}
Harsha Nori, Samuel Jenkins, Paul Koch, and Rich Caruana.
\newblock Interpretml: A unified framework for machine learning
  interpretability.
\newblock \emph{arXiv preprint arXiv:1909.09223}, 2019.

\bibitem[Pace and Barry(1997)]{pace1997sparse}
R~Kelley Pace and Ronald Barry.
\newblock Sparse spatial autoregressions.
\newblock \emph{Statistics \& Probability Letters}, 1997.

\bibitem[Pedregosa et~al.(2011)Pedregosa, Varoquaux, Gramfort, Michel, Thirion,
  Grisel, Blondel, Prettenhofer, Weiss, Dubourg, et~al.]{pedregosa2011scikit}
Fabian Pedregosa, Ga{\"e}l Varoquaux, Alexandre Gramfort, Vincent Michel,
  Bertrand Thirion, Olivier Grisel, Mathieu Blondel, Peter Prettenhofer, Ron
  Weiss, Vincent Dubourg, et~al.
\newblock Scikit-learn: Machine learning in python.
\newblock \emph{JMLR}, 2011.

\bibitem[Potts(1999)]{potts1999generalized}
William~JE Potts.
\newblock Generalized additive neural networks.
\newblock \emph{SIGKDD}, 1999.

\bibitem[ProPublica(2016)]{compas}
ProPublica.
\newblock {COMPAS Data and analysis for `Machine Bias'}.
\newblock \url{https://github.com/propublica/compas-analysis}, 2016.

\bibitem[Radford et~al.(2018)Radford, Wu, Child, Luan, Amodei, and
  Sutskever]{radford2019language}
Alec Radford, Jeffrey Wu, Rewon Child, David Luan, Dario Amodei, and Ilya
  Sutskever.
\newblock Language models are unsupervised multitask learners.
\newblock 2018.

\bibitem[Rahaman et~al.(2018)Rahaman, Baratin, Arpit, Draxler, Lin, Hamprecht,
  Bengio, and Courville]{rahaman2018spectral}
Nasim Rahaman, Aristide Baratin, Devansh Arpit, Felix Draxler, Min Lin, Fred~A
  Hamprecht, Yoshua Bengio, and Aaron Courville.
\newblock On the spectral bias of neural networks.
\newblock \emph{ICML}, 2018.

\bibitem[Ribeiro et~al.(2016)Ribeiro, Singh, and Guestrin]{ribeiro2016should}
Marco~Tulio Ribeiro, Sameer Singh, and Carlos Guestrin.
\newblock " why should i trust you?" explaining the predictions of any
  classifier.
\newblock \emph{SIGKDD}, 2016.

\bibitem[Roscher et~al.(2019)Roscher, Bohn, Duarte, and
  Garcke]{roscher2019explainable}
Ribana Roscher, Bastian Bohn, Marco~F Duarte, and Jochen Garcke.
\newblock Explainable machine learning for scientific insights and discoveries.
\newblock \emph{arXiv preprint arXiv:1905.08883}, 2019.

\bibitem[Rudin(2019)]{rudin2019stop}
Cynthia Rudin.
\newblock Stop explaining black box machine learning models for high stakes
  decisions and use interpretable models instead.
\newblock \emph{Nature Machine Intelligence}, 2019.

\bibitem[Rudin et~al.(2021)Rudin, Chen, Chen, Huang, Semenova, and
  Zhong]{rudin2021interpretable}
Cynthia Rudin, Chaofan Chen, Zhi Chen, Haiyang Huang, Lesia Semenova, and Chudi
  Zhong.
\newblock Interpretable machine learning: Fundamental principles and 10 grand
  challenges.
\newblock \emph{arXiv preprint arXiv:2103.11251}, 2021.

\bibitem[Rumelhart et~al.(1986)Rumelhart, Hinton, and
  Williams]{rumelhart1986learning}
David~E Rumelhart, Geoffrey~E Hinton, and Ronald~J Williams.
\newblock Learning representations by back-propagating errors.
\newblock \emph{Nature}, 1986.

\bibitem[Saeed et~al.(2011)Saeed, Villarroel, Reisner, Clifford, Lehman, Moody,
  Heldt, Kyaw, Moody, and Mark]{saeed2011multiparameter}
Mohammed Saeed, Mauricio Villarroel, Andrew~T Reisner, Gari Clifford, Li-Wei
  Lehman, George Moody, Thomas Heldt, Tin~H Kyaw, Benjamin Moody, and Roger~G
  Mark.
\newblock Multiparameter intelligent monitoring in intensive care ii
  (mimic-ii): a public-access intensive care unit database.
\newblock \emph{Critical care medicine}, 2011.

\bibitem[Snoek et~al.(2012)Snoek, Larochelle, and Adams]{snoek2012practical}
Jasper Snoek, Hugo Larochelle, and Ryan~P Adams.
\newblock Practical bayesian optimization of machine learning algorithms.
\newblock \emph{NeurIPS}, 2012.

\bibitem[Srivastava et~al.(2021)Srivastava, Aydin, Mazzoni, and
  Mahony]{srivastava2021interpretable}
Divyanshi Srivastava, Beg{\"u}m Aydin, Esteban~O Mazzoni, and Shaun Mahony.
\newblock An interpretable bimodal neural network characterizes the sequence
  and preexisting chromatin predictors of induced transcription factor binding.
\newblock \emph{Genome Biology}, 2021.

\bibitem[Srivastava et~al.(2014)Srivastava, Hinton, Krizhevsky, Sutskever, and
  Salakhutdinov]{srivastava2014dropout}
Nitish Srivastava, Geoffrey Hinton, Alex Krizhevsky, Ilya Sutskever, and Ruslan
  Salakhutdinov.
\newblock Dropout: a simple way to prevent neural networks from overfitting.
\newblock \emph{JMLR}, 2014.

\bibitem[Tan et~al.(2018)Tan, Caruana, Hooker, and Lou]{tan2018distill}
Sarah Tan, Rich Caruana, Giles Hooker, and Yin Lou.
\newblock Distill-and-compare: Auditing black-box models using transparent
  model distillation.
\newblock \emph{{AAAI/ACM Conference on AI, Ethics, and Society}}, 2018.

\bibitem[Tancik et~al.(2020)Tancik, Srinivasan, Mildenhall, Fridovich-Keil,
  Raghavan, Singhal, Ramamoorthi, Barron, and Ng]{tancik2020fourier}
Matthew Tancik, Pratul~P Srinivasan, Ben Mildenhall, Sara Fridovich-Keil,
  Nithin Raghavan, Utkarsh Singhal, Ravi Ramamoorthi, Jonathan~T Barron, and
  Ren Ng.
\newblock Fourier features let networks learn high frequency functions in low
  dimensional domains.
\newblock \emph{arXiv preprint arXiv:2006.10739}, 2020.

\bibitem[{The Royal Society}(2019)]{explainable_royalsociety}
{The Royal Society}.
\newblock {Explainable AI: The Basics - Policy Briefing}.
\newblock 2019.
\newblock
  \url{https://royalsociety.org/-/media/policy/projects/explainable-ai/AI-and-interpretability-policy-briefing.pdf}.

\bibitem[{The Royal Society and The Alan Turing
  Institute}(2019)]{ai_in_science}
{The Royal Society and The Alan Turing Institute}.
\newblock {The AI revolution in scientific research}.
\newblock 2019.
\newblock
  \url{https://royalsociety.org/-/media/policy/projects/ai-and-society/AI-revolution-in-science.pdf}.

\bibitem[Utkin et~al.(2021)Utkin, Satyukov, and Konstantinov]{utkin2021survnam}
Lev~V Utkin, Egor~D Satyukov, and Andrei~V Konstantinov.
\newblock Survnam: The machine learning survival model explanation.
\newblock \emph{arXiv preprint arXiv:2104.08903}, 2021.

\end{thebibliography}
